\documentclass[10pt,twocolumn,letterpaper]{article}

\usepackage[pagenumbers]{cvpr} 

\usepackage{comment}
\usepackage{amsmath,amssymb} 
\usepackage{times}

\usepackage[dvipsnames]{xcolor}

\usepackage{breqn}
\usepackage{amsmath}

\usepackage[dvipsnames]{xcolor}
\usepackage{gensymb}
\usepackage[]{multirow}
\usepackage{graphbox}
\usepackage{epsfig}
\usepackage{graphicx}
\usepackage{amsmath}
\usepackage{amssymb}
\usepackage{enumitem}
\usepackage[]{multirow}
\usepackage[]{booktabs}
\usepackage{amsfonts}
\usepackage[accsupp]{axessibility}
\usepackage{mwe}
\usepackage{graphbox}
\usepackage{lipsum}
\usepackage{tabularx}
\usepackage{bm}
\usepackage{overpic}

\newcommand{\rone}[1]{\textcolor{blue}{R1}} 
\newcommand{\rtwo}[1]{\textcolor{magenta}{R2}} 
\newcommand{\rthree}[1]{\textcolor{teal}{R3}} 

\newcommand{\name}{ANIM\xspace}
\newcommand{\datasetname}{ANIM-Real\xspace}

\definecolor{cvprblue}{rgb}{0.21,0.49,0.74}
\usepackage[pagebackref,breaklinks,colorlinks,citecolor=cvprblue]{hyperref}

\def\paperID{16047} 
\def\confName{CVPR}
\def\confYear{2024}

\title{ANIM: Accurate Neural Implicit Model \\for Human Reconstruction from a single RGB-D image}

\author{Marco Pesavento\textsuperscript{\rm 1,3\thanks{Work performed during an internship at Meta Reality Labs, Sausalito}} \and Yuanlu Xu\textsuperscript{\rm 3} \and Nikolaos Sarafianos\textsuperscript{\rm 3}  \and Robert Maier\textsuperscript{\rm 3}  \and Ziyan Wang\textsuperscript{\rm 3} \and Chun-Han Yao\textsuperscript{\rm 2}  \and Marco Volino\textsuperscript{\rm 1} \and Edmond Boyer\textsuperscript{\rm 3} \and Adrian Hilton\textsuperscript{\rm 1} \and Tony Tung\textsuperscript{\rm 3}
\\
\vspace{-5mm}
\and \textsuperscript{\rm 1}University of Surrey, CVSSP, UK \and \textsuperscript{\rm 2}UC Merced \and \textsuperscript{\rm 3}Meta Reality Labs 
}
\begin{document}
\maketitle
\vspace{-6mm}
\begin{abstract}
\vspace{-2mm}
Recent progress in human shape learning, shows that neural implicit models are effective in generating 3D human surfaces from limited number of views, and even from a single RGB image. 
However, existing monocular approaches still struggle to recover fine geometric details such as face, hands or cloth wrinkles. They are also easily prone to depth ambiguities that  result in distorted geometries along the camera optical axis.
In this paper, we explore the benefits of incorporating depth observations in the reconstruction process by introducing \name, a novel method that reconstructs arbitrary 3D human shapes from single-view RGB-D images with an unprecedented level of accuracy.
Our model learns geometric details from both multi-resolution pixel-aligned and voxel-aligned features to leverage depth information and enable spatial relationships, mitigating depth ambiguities.
We further enhance the quality of the reconstructed shape by 
introducing a depth-supervision strategy, which improves the accuracy of the signed distance field estimation of points that lie on the reconstructed surface.
Experiments demonstrate that \name outperforms state-of-the-art works that use RGB, surface normals, point cloud or RGB-D data as input.
In addition, we introduce \datasetname, a new multi-modal dataset comprising high-quality scans paired with consumer-grade RGB-D camera, and our
protocol to fine-tune \name, enabling high-quality reconstruction from real-world human capture. \href{https://marcopesavento.github.io/ANIM/}{https://marcopesavento.github.io/ANIM/}
\end{abstract}
\begin{figure}[ptb]
\vspace{-4mm}
\centering
\includegraphics[width=0.95\linewidth]{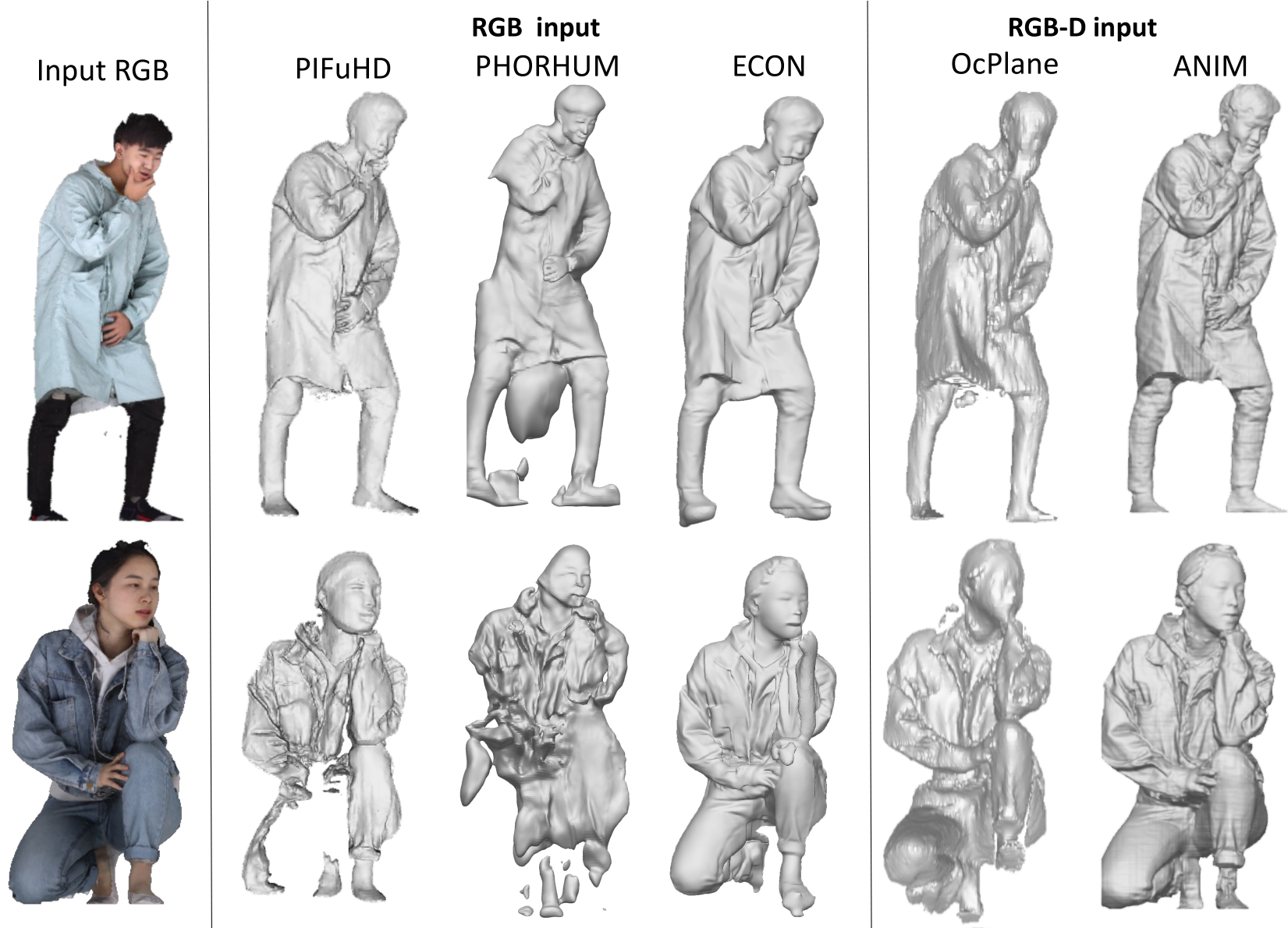}
\vspace{-4mm}
\caption{\name enables human shape reconstruction with higher accuracy and without shape distortions compared to the state-of-the-art methods based on monocular RGB-D or RGB input.}
\label{fig:teaser}
\vspace{-7mm}
\end{figure}

\vspace{-4mm}
\section{Introduction}
\vspace{-1mm}
\label{sec:intro}
The increasing interest in 3D virtual world creation has led to a substantial demand for easily accessible 3D reconstruction solutions.
Consequently, this has emerged as a prominent research domain in computer vision with applications in virtual and augmented reality, gaming, medicine and e-shopping, among others. 
A recurrent challenge revolves around ensuring the fidelity of the created models and, consequently, the accuracy of the reconstruction methods, especially when reconstructing 3D avatars of real people.
To this aim, depth sensors that are nowadays ubiquitous in commercial devices (\eg, LiDAR Depth Camera, AI Stereo Depth Map, Azure Kinect, and Asus) can be leveraged to develop  efficient and accurate reconstruction solutions.
Our objective is to build high-fidelity models of clothed humans from single RGB-D images by learning an Accurate Neural Implicit Model (\name).
Monocular approaches based on generative adversarial networks~\cite{wang2020normalgan,li2022sfnet} produce realistic front and back depths of the model. However, their fidelity is  limited as the prediction depends on the generative ability of the network.
While leveraging priors such as parametric body models can produce complete body shapes~\cite{lu20223d}, they often lack details.
Several works resort to multiple RGB-D images or monocular videos, combining multiple predictions to reconstruct higher-quality 3D shapes~\cite{dong2021geometry,dong2022pina,su2022robustfusion}.
In contrast to previous works that exclusively process either single RGB images~\cite{PIFuICCV19,SuRSECCV2022,saito2020pifuhd,ICONCVPR2022, Zerong2020PaMIR} or 3D point clouds~\cite{chibane20ifnet,tiwari21neuralgif, bharat20ipnet}, our proposed approach, which relies  on a neural implicit model (a learned Signed Distance Field, or SDF), reconstructs accurate 3D models of clothed humans from a single RGB-D image with significantly higher levels of detail.
Related approaches that also estimate an implicit representation of the 3D shape from a single RGB-D image either lack pixel-alignment with the input, as in  OPlanes~\cite{zhao2022occupancy}, or estimate depth from the RGB input, limiting therefore the fidelity, and rely on 3D parametric models, as DiFU \cite{Song2022difu}.
Our representation is a pixel-voxel-aligned implicit model of the reconstructed surface learned using a combination of multi-resolution 2D feature extractor and a specific SparseConvNet U-Net (the Volume Feature Extractor or VFE) to process multi-resolution 2D and volumetric features. A depth-supervision strategy is also introduced to further enhance the SDF estimation.
We demonstrate the advantages of using RGB-D images over alternative methods that suffer from depth ambiguity or that reproduce low-fidelity details, as illustrated in Figure~\ref{fig:teaser}.
Our extensive experiments show that \name outperforms existing methods that reconstruct 3D human shapes from single RGB images, surface normals, point cloud or RGB-D data.
In practice, consumer-grade RGB-D cameras produce noisier data compared to high-end 3D scanners. This directly impacts the 3D reconstruction quality since the surface estimation builds on features learned from the input. In order to reduce this impact and to achieve high-fidelity reconstruction  with consumer-grade camera, we propose to learn a model trained with high-quality 3D ground-truth data paired with real noisy RGB-D data as input.
Since datasets with these characteristics are currently unavailable, we introduce the multi-modal dataset \datasetname which includes 3D scans reconstructed from a high-resolution multi-view camera system aligned with RGB-D data captured by a consumer-grade camera.
Fine-tuning \name with \datasetname enables to better handle sensor noise and to obtain high-quality 3D shape models from real-world capture.

\noindent In summary, our contributions are:
\begin{itemize}
\item A novel network architecture \name that includes a pixel-voxel-aligned implicit representation obtained from the 3D Volume Feature Extractor and 2D multi-resolution features to reconstruct accurate and high-fidelity 3D human shape from a single-view RGB-D image.
\item A novel depth-supervision strategy that refines the SDF learning of the 3D points lying on the reconstructed surface by leveraging the input point cloud.
\item The multi-modal dataset \datasetname comprising synchronous captures from a high-quality 3D human scanner aligned with a consumer-grade RGB-D camera, and a protocol to fine-tune \name real-world human capture.
\item Unprecedented quantitative and qualitative results for human shape reconstruction from single RGB-D images. 
\end{itemize}
\begin{figure*}[ptb]
\centering
\vspace{-3mm}
\includegraphics[width=0.9\textwidth]{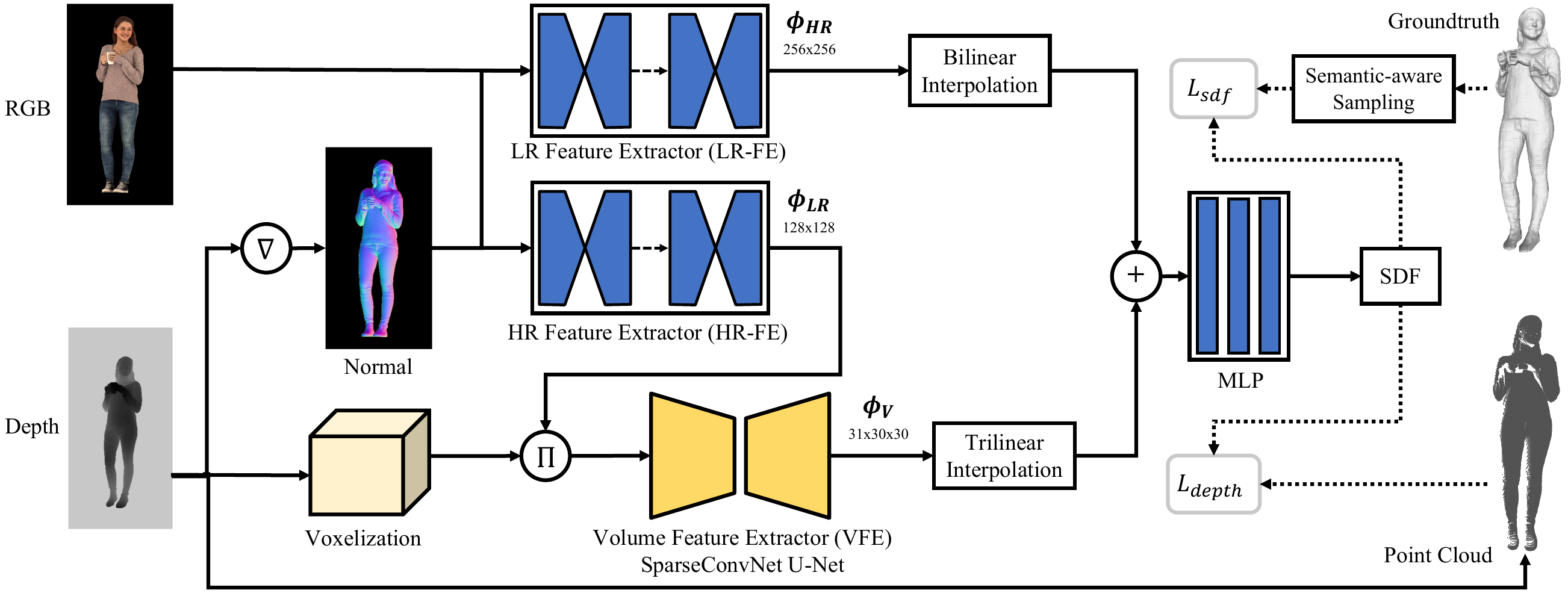}
\vspace{-4mm}
\caption{\name architecture. Our proposed framework has three major components: i) a multi-resolution appearance feature extractor for color and normal inputs (LR-FE and HR-FE), ii) a novel SparseConvNet U-Net (Volume Feature Extractor or VFE) that efficiently extracts geometry features from 3D voxels and low-resolution image features, iii) an MLP that estimate the implicit surface representation of full-body humans. $+$ denotes concatenation, $\Pi$ means fetching pixel-aligned 2D LR features and concatenating with 3D voxels, and $\nabla$ indicates gradient operation applied to retrieve normals from depth map (using neighboring pixel cross-product).
}
\vspace{-6mm}
\label{fig:framework}
\end{figure*}
%
%
\vspace{-1mm}
\section{Related Work}
\label{sec:relatedwork}
\vspace{-1mm}
%
%
%
%
%
\textbf{Reconstruction from single-view images}. 
Single-view 3D human reconstruction has been approached using a wide range of methods and representations. 
Representations used in this domain include voxels~\cite{varol18_bodynet, VolumeRegECCVW2018,zheng2019deephuman}, two-way depth maps~\cite{gabeur2019moulding, smith2019facsimile}, visual hull~\cite{SiCloPeCVPR19}, parametric models~\cite{loper2015smpl,bogo2016keep,kanazawa2018end,pavlakos2019expressive,kocabas2020vibe,xu20213d,alldieck2018video,alldieck2019tex2shape,alldieck2019learning}. 
These methods cannot reproduce high-quality 3D human shapes, with only minimally clothed reconstruction. %
In contrast, implicit function representations have shown great promise for the task of human digitization from a single image~\cite{PIFuICCV19, saito2020pifuhd, he2020geopifu,huang2020arch,he2021arch++,SuRSECCV2022,li2020robust, li2020monocular}. 
One of the first approaches to adopt this representation was PIFu~\cite{PIFuICCV19}, which exploits pixel-aligned image features rather than global features to preserve local details of the input image as the occupancy of any 3D point is predicted. 
SuRS \cite{SuRSECCV2022} demonstrates fine-scale detail can be recovered even from low-resolution input images using a super-resolution learning framework.
PaMIR~\cite{Zerong2020PaMIR} concatenates 3D features extracted from an estimated SMPL model with 2D features. 
In PIFuHD~\cite{saito2020pifuhd}, the quality of the reconstruction is improved by using surface normals and a coarse-to-fine implicit function framework.
Alldieck \etal~\cite{alldieck2022photorealistic} improved upon PIFuHD by estimating the 3D geometry, the surface albedo, and shading, from a single image in a joint manner.
ICON~\cite{ICONCVPR2022} improves the estimation of front and back normals used for the reconstruction by guiding it with a parametric model whilst ECON~\cite{xiu2023econ} addresses the problem of clothing reconstruction of the SMPL body by feeding the estimated normals into a d-BiNI optimizer.
%
%
%
\\\noindent \textbf{Reconstruction from RGB-D and point clouds}.
Works that use RGB-D images as input for the task of clothed human reconstruction fall into two categories taking a single RGB-D image \cite{wang2020normalgan, zhao2022occupancy} or multiple sequences of RGB-D images \cite{dong2022pina,burov2021dynsurfun,li2022sfnet,mao2017easyfastrecon} as input.
Methods that consider RGB-D sequences have to fuse multiple partial and noisy observations into a coherent model. 
In contrast, single view RGB-D methods have to tackle the problem of shape completion \cite{li2017tvcg,zhang2018deepdepth} due to partial observations leading to incomplete reconstructions. Approaches that estimate an implicit representation from a single RGB-D image cannot achieve the same level of quality and accuracy as the proposed method~\cite{Song2022difu,zhao2022occupancy}.
DiFU \cite{Song2022difu} estimates the implicit representation by using a SMPL~\cite{loper2015smpl} voxel encoder, an U-Net depth estimator, and a scale regressor. The back and front depth maps are estimated from the RGB image, introducing noise and limiting the quality of the reconstructed shapes. OcPlanes~\cite{zhao2022occupancy} adopts a plane-aligned occupancy function to align the feature extracted from the input image to the input depth. Replacing the local pixel-alignment with a global alignment reduces the quality of the reconstruction.
\\Point clouds are an alternative representation explored for the task of 3D human reconstruction\cite{POP:ICCV:2021,SkiRT:3DV:2022,chibane20ifnet,bhatnagar2020ipnet}.
IF-Net~\cite{chibane20ifnet} exploits partial point clouds and learns implicit functions using latent voxel features.  IP-Net \cite{bhatnagar2020ipnet} further develops this idea by incorporating SMPL\cite{loper2015smpl} into the pipeline to enable animatable reconstructions.
In comparison to prior work, \name reconstructs 3D human shapes from single-view RGB-D images at an unprecedented level of detail and reconstruction accuracy via the proposed novel architecture and depth-supervision strategy. 

\noindent \textbf{Single-view RGB-D datasets for human reconstruction}.
Available datasets containing RGB-D data from consumer-grade cameras are primarily designed for different tasks such as people re-identification~\cite{barbosa2012re,munaro2014one} or human activity recognition~\cite{Coppola2016a,xia2012view,liu2017pku,gaglio2014human,wolf2014evaluation}. 3D ground-truth shapes are not provided in these datasets since the body skeleton is sufficient to achieve the intended tasks. Human3.6m~\cite{ionescu2014human3} provides 3D human shape as ground truth along with depth maps.
However, the quality of the shapes is limited, lacking clothing and details of the human body. Training neural implicit models with this dataset restricts the ability to learn high-fidelity clothed 3D human shapes. 
Recently, SynWild~\cite{guo2023vid2avatar} used RGB and IR cameras to create the 3D ground truth but the semi-synthetic dataset is created by rendering the monocular video with a virtual camera, which is not affected by real-world noise.
We propose the novel multi-modal dataset \datasetname that includes high-quality 3D human shapes reconstructed from a multi-view camera system, aligned and synchronized with real-world RGB-D data acquired with a consumer-grade camera.
\vspace{-1mm}
\section{Methodology}
\label{sec:method}
\vspace{-1mm}
\name learns an implicit function $f$ to reconstruct accurate and high-fidelity human shapes from a single RGB-D image.
We present an end-to-end framework that takes an RGB-D image as input and estimates the SDF of the person. 
Specifically, as illustrated in~\cref{fig:framework}, \name extracts a high-resolution (HR) 2D feature to encode high-frequency details and a low-resolution (LR) feature to maintain holistic reasoning from a concatenation of the input colour and normal, considering their shared image-space properties. The low-resolution features serve as a prior for a novel SparseConvNet~\cite{SparseConvNet18} U-Net, which extracts geometric features by processing 3D voxels created from the depth map and concatenated with its low-resolution image-space features. Given appearance and geometry features, an MLP predicts the SDF of the reconstructed subject. We train the framework end-to-end with a novel depth-supervision strategy that refines the estimation of the SDF of the 3D points close to the reconstructed surface by leveraging the input point cloud. 
Compared with related methods, our approach fuses information across multiple modes and is thus more robust to depth ambiguity and challenging poses.

\vspace{-1mm}
\subsection{Accurate Implicit Surface Estimation}
\vspace{-1mm}

Assuming the 3D clothed human to be reconstructed as a one-layer watertight mesh, we represent it with an implicit surface function $f$. The value $f(x)$ of a point $x\in\mathbb{R}^3$ denotes the distance of this point to its closest surface. To obtain a surface, we can simply threshold $f$ to obtain an isosurface $f(x) = \tau$. The surface to be reconstructed is then defined as the zero level-set of $f$:
\begin{equation}\small
    f' = \{ x:\; f(x) = 0,\; x\in\mathbb{R}^3\}.
\end{equation}
Fine surface details are stored in high frequency and need to be represented on the final shape, which has to be robust to depth ambiguity. 
Recent approaches show the effect of representing the shape with an implicit function aligned with the input data. PIFu~\cite{PIFuICCV19} introduced the concept of pixel-alignment to increase the quality of 3D human shapes by projecting a 3D point $x \in \mathbb{R}^3 $ in the image feature $\phi(I)$ of an input RGB image $I$. PaMIR~\cite{Zerong2020PaMIR}  proposes a voxel-aligned implicit function to leverage spatial information from a parametric model to avoid depth ambiguity. We propose a novel architecture that learns a high-fidelity implicit surface representation $s_{HF}$ that is both pixel-aligned with the input image $I$ and surface normal $S_N$ and voxel-aligned with the voxel created from the input depth map $D$:
\vspace{-5mm}
\begin{equation}\small
\hat{s}_{HF}=f_x\Bigl(\phi_{HR}\bigl(\pi(I,S_N)\bigr),z(x),\gamma(D)\Bigr), \hat{s}_{HF} \in \mathbb{R},
\vspace{-1mm}
\end{equation}
where $\phi_{HR}(I,S_N)$ are the HR features extracted from the concatenation between the input image $I$ and the surface normal $S_N$, $\pi$ is the orthographic projection, $\gamma(D)=\phi_{V}(\phi_{LR}(I,S_N),D)$ is the feature extracted from the depth $D$ linked with LR feature $\phi_{LR}$ retrieved from $I$, and $S_N$. $z(x)$ is the depth value of $x$. To estimate the implicit representation $\hat{s}_{HF}$, \name comprises the following modules. 
\\\noindent\textbf{2D Feature Extractor.} 2D pixel-aligned features are obtained from the input image $I$ and surface normals $S_N$. LR features $\phi_{LR}(I,S_N)$ are extracted with a stacked hourglass network (LR-FE) to guarantee a large receptive field, which is required to maintain holistic reasoning~\cite{PIFuICCV19}. LR features are also used by the 3D feature extractor since the SparseConvNet requires that each input voxel is linked to an embedding. Instead of using random embedding like previous works~\cite{peng2020neural}, performance improves when features extracted from the input image are used (see~\cref{ssec:abl}). High-quality details cannot be reconstructed if only LR features are used. To embed local details of $I$ and $S_N$, we introduce a second stacked hourglass architecture (HR-FE) to retrieve HR features $\phi_{HR}(I,S_N)$. These HR features are pixel-aligned with the input data via orthographic projection $\pi$.
\\\noindent\textbf{3D Feature Extractor.} Learning spatial relationships in 3D space is fundamental to solving the problems of depth ambiguity derived by the single-view input. We process a voxel retrieved from the input depth $D$ with a novel SparseConvNet U-Net style architecture: Voxel Feature Extractor (VFE). Due to the requirements of the VFE, the LR feature $\phi_{LR}(I,S_N)$ are linked to the voxels to provide additional information from the 2D RGB input before extracting the 3D features.
Instead of using a 3D convolutional neural network as in~\cite{Zerong2020PaMIR,he2020geopifu}, we extract voxel-aligned features $\gamma(D)$ with 3D sparse convolution layers, which have been proven to be efficient when the input is sparse such as in the point cloud created from a single view.  This also ensures a performance gain at training and testing times with faster speed in the order of magnitude compared to the 3D ConvNets. Voxel-alignment is obtained by trilinear interpolation of 3D points $x$ with $\phi_V$. 
\\\noindent\textbf{Multi-layer Perceptron (MLP).} The 2D pixel-aligned features are concatenated with the 3D voxel-aligned feature and processed by a multi-layer perceptron that models the implicit function $f_x$ and estimates the final SDF $\hat{s}_{HF}$.

\vspace{-1mm}
\subsection{Depth-Supervision Strategy}
\vspace{-1mm}

To improve the learning of the SDF of the 3D reconstructed surface, we propose to leverage the depth channel of the RGB-D input and estimate an implicit representation of the input sparse point cloud $\zeta$ by extracting pixel-aligned feature $\phi_{HR}(I,S_N)$ from HR-FE and voxel-aligned features
$\gamma(\zeta)=\phi_{V}(\phi_{LR}(I,S_N),\zeta)$ from the VFE. An implicit function $f_{\zeta}$  representing $\zeta$ is learned with an MLP that shares the weight with the one previously applied:
\begin{equation}\small
    \hat{s}_{\zeta}=f_{\zeta}\Bigl(\phi_{HR}\bigl(\pi(I,S_N)\bigr),z(x_{\zeta}),\gamma(\zeta)\Bigr), \;\;\; \hat{s}_{\zeta} \in \mathbb{R},
\end{equation}
where $x_{\zeta}$ are the points of the input point cloud, which are projected into $\phi_{HR}(I,S_N)$ for pixel alignment. The SDF of $x_{\zeta}$ should be $0$ since $x_{\zeta}$ lies on the surface of the reconstructed shape. The network significantly improves its ability to estimate which points lie on the surface (\cref{ssec:abl}).

The network is trained end-to-end with two Huber losses, one to train the implicit function $f_x$ of 3D points $x$ sampled on the 3D ground-truth shape: 
\begin{equation}\small
\begin{aligned}
L_{sdf} = \begin{cases} 0.5\,({\hat{s}_{HF}-s_{HF}})^2, \small{\quad \text{if $\|\hat{s}_{HF}-s_{HF}\|_2< \delta $}},  \\ \delta\,(|\hat{s}_{HF}-s_{HF}|-0.5\delta), \quad\text{otherwise}, \end{cases}
\end{aligned}
\end{equation}
and the other for the depth-supervision strategy: 
\begin{equation}\small
\begin{aligned}
L_{depth} = \begin{cases} 0.5\,({\hat{s}_{\zeta}-s_{\zeta}})^2, \quad\text{if $\|\hat{s}_{\zeta}-s_{\zeta}\|_2< \delta $},  \\ \delta\,(|\hat{s}_{\zeta}-s_{\zeta}|-0.5\delta), \quad\text{otherwise}, \end{cases}
\end{aligned}
\end{equation}
where $s_{\zeta} $ is the ground-truth label for the $N_{\zeta}$ points of the depth map $\zeta$ and $\delta$ is a threshold for estimation correctness.
\\The learning is thus supervised with both points sampled in the 3D space of the full body to learn its full representation, and with points from the depth to improve the SDF estimation of points that lie on the visible surface, resulting in a more accurate representation of high-quality details.

\begin{figure}[ptb]
\vspace{-3mm}
\centering
\includegraphics[width=0.85\linewidth]{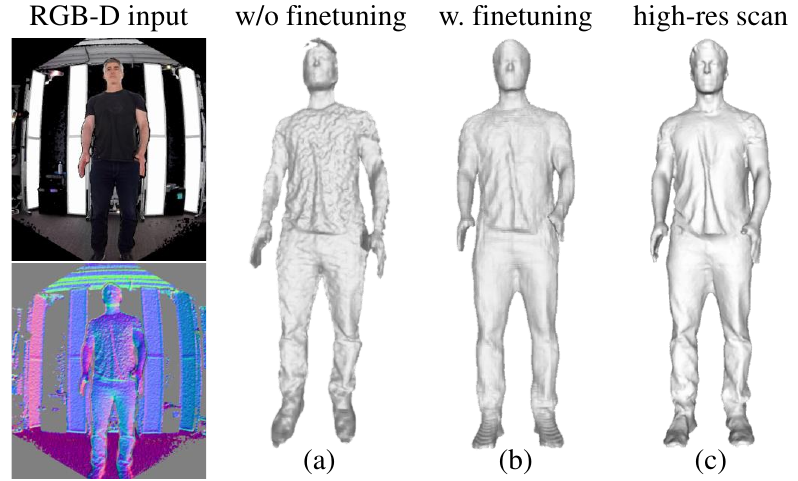} 
\vspace{-3mm}
\caption{\name reconstructions from real-world capture with a consumer-grade RGB-D camera (Azure Kinect) before (a), and after (b) fine-tuning on the proposed dataset \datasetname, which quality is closer to a high-res scan capture (c).}
\label{fig:noise}
\vspace{-5mm}
\end{figure}

\noindent\textbf{Inference.} Instead of using $M^3$ random 3D points distributed in a 3D grid in space as related works, we consider the input point cloud to create a 3D grid to sample the SDF. A bounding box is created around the sparse point cloud augmented with Gaussian sampling. The resolution of the novel 3D grid is computed as ($m\!\times\! H, m\!\times\! W, m\!\times\! D$) where $m=\sqrt[3]{M^3/L}$, $M=256$, $L=H\!\times\! W\!\times\! D$ and $(H, W, D)$ is the dimension of the bounding box. The points are not randomly distributed in a squared grid but are concentrated in the region where the shape is generated. The 2D and 3D features are extracted from the RGB-D input and aligned with the grid of points. The final SDF is estimated with the MLP and the shape is obtained by extracting iso-surface $f_x=0$ of the probability field $\hat{s}_{HF}$ via Marching Cubes~\cite{lorensen1987marching}.
\\\noindent\textbf{Extension to consumer-grade RGB-D camera.}
RGB-D data acquired in real-world scenarios are affected by the noise propagated in the capture systems.
Applying \name directly on data acquired with a consumer-grade camera (\eg Azure Kinect) reveals severe reconstruction artifacts due to the sensor noise (\cref{fig:noise}a).
\name should learn the noise added to captured data to perform effectively in real-world scenarios. A solution is to fine-tune \name with real noisy data.
However, datasets that combine RGB-D data with high-resolution 3D ground-truth shapes, necessary to achieve high-quality reconstruction, are unavailable.
We thus create the new dataset \datasetname with a multi-modal setup consisting of a consumer-grade RGB-D camera and a high-resolution 4D scanner, as explained in~\cref{sec:dataset}. We then fine-tune \name parameters on \datasetname to learn typical consumer-grade sensor noise, resulting in a significant improvement in the 3D shape (see~\cref{fig:noise}b).

\vspace{-0.15cm}
\section{Datasets}
\label{sec:dataset}
\vspace{-1mm}

\noindent\textbf{\datasetname dataset}. 
We present \datasetname, a new multi-modal dataset that can be used to obtain high-quality reconstructions from consumer-grade RGB-D camera for real-world applications.
Consumer-grade monocular RGB-D cameras (\eg Azure Kinect) while ubiquitous produce less accurate and incomplete reconstructions than high-end capture systems such as full-body 3D scanners that are based on multi-view capture~\cite{3dmd}.
Depth sensors from consumer-grade cameras contain noise that directly affects surface reconstruction accuracy~\cite{depth2019,depth2022}. 
To date, it is not possible to generate high-fidelity full-body 3D reconstruction from a one-shot capture with a consumer-grade RGB-D sensor.
Hence, we create \datasetname as follows:
\begin{itemize}
\item We acquire high-quality 3D scans with a high-resolution camera system that uses active stereo and multi-view cameras~\cite{3dmd}. It comprises 16 high-resolution RGB cameras and stereo pairs. However, raw scan reconstructions can contain holes from self-occlusions, lack of coverage, or challenging regions (\eg hair), leading to incorrect sampling and SDF estimation.
We therefore apply the Fast Winding Numbers algorithm~\cite{Barill:FW:2018}, hence producing high-quality watertight shapes (see~Fig.\ref{fig:noise}c).
\item RGB-D data is captured with a consumer-grade camera (Azure Kinect), to allow \name to learn the sensor noise introduced in the input RGB-D images (see~Fig.\ref{fig:noise}).
\item Intrinsics and extrinsics camera calibrations and capture synchronization are crucial to align the 3D shape with the corresponding RGB-D input. 
The extrinsics calibration between capture systems is obtained using a generic calibration object with salient shapes, while synchronization is obtained using Sync I/O and generated trigger signals. 
The transformation matrix obtained with the calibration is used to project the 3D points sampled in the 3D shape to the 2D image feature, achieving pixel-alignment. 
The input depth map is also aligned with the 3D ground-truth shape, ensuring voxel-alignment.
\item For evaluation, we simultaneously capture 28 subjects in motion using the 2 systems. 
We fine-tune \name with an additional 16k frames, consisting of around 800 frames on average from a single view of 21 subjects.
\end{itemize}


\noindent\textbf{Synthetic datasets}. For additional quantitative and qualitative evaluations, we use large public synthetic datasets of 3D humans in various poses and clothing, providing a comprehensive assessment of the effectiveness of the proposed approach.
We use 909 RenderPeople~\cite{renderpeople} scans and split them into 800 for training and 109 for testing. To evaluate the generalization power of \name, we use 200 human models THuman2.0 dataset~\cite{tao2021function4d} as another test set. We evaluate the reconstruction accuracy with 3 quantitative metrics: the average point-to-surface Euclidean distance (P2S), the normal reprojection~\cite{PIFuICCV19}, and the Chamfer distance (CD), cm.
\vspace{-3mm}
\section{Experiments} \label{sec:experiments}
\vspace{-1mm}

We quantitatively and qualitatively evaluate the proposed approach to the task of reconstructing 3D human shapes from a single RGB-D image. We conduct in-depth ablation studies where the proposed modules are removed. Finally, we show reconstruction examples from real-world RGB-D data where \name faithfully reconstructs the clothed geometry despite the presence of sensor noise. 

\begin{figure}[ptb]
\vspace{-3mm}
\centering
\includegraphics[width=0.9\linewidth]{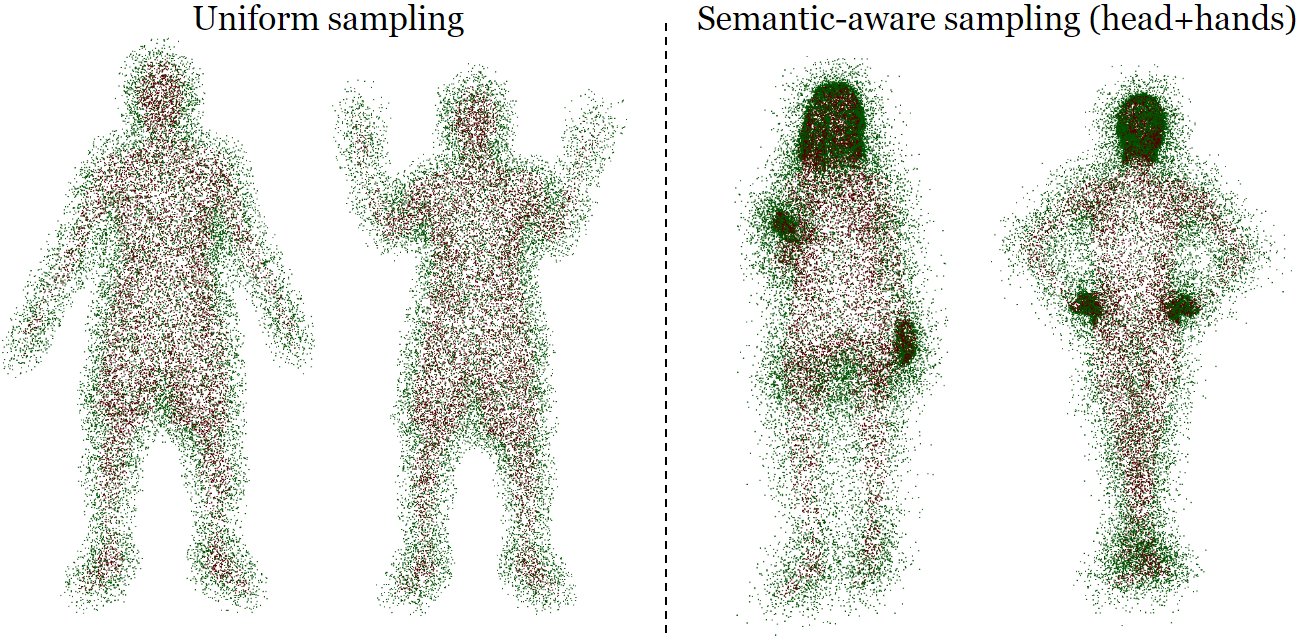}
\vspace{-4mm}
\caption{Semantic-aware sampling. Compared to uniform sampling (left), semantic-aware sampling (right) enables finer learning of human features on specific regions such as the head and hands.}
\label{fig:marco_sampling}
\vspace{-5mm}
\end{figure}
\subsection{Implementation details}
We render the training RGB-D images and surface normals at 512x512 pixel resolution with a virtual camera that rotates around the 3D model with a step of \(2^o\). 
\\To create the ground-truth SDF $s_{HF}$, state-of-the-art works~\cite{PIFuICCV19, saito2020pifuhd,Zerong2020PaMIR} label a set of 3D points that are sampled around the surface with a mixture of uniform sampling and randomly added offset following normal distribution $\mathcal{N}(0,\sigma_{LR})$. 
Since the sampling is homogeneous, smaller parts of the body have a lower number of 3D sampled points and cannot be reconstructed by these methods. We propose to augment the sampling in a semantic-aware manner.
\\\textbf{Semantic-Aware Body Surface Sampling.} We augment the number of points sampled in the face and hands regions by selecting the points that reproject in the same regions of a semantic mask, estimated from the RGB image using body-part segmentation~\cite{ianina2022bodymap}. We project $X_{b}=48,000$ 3D points sampled around the ground-truth surface on the 2D mask. If the point is projected onto the face or hand regions of the mask, we sample additional points around that point:
\vspace{-3mm}
\begin{equation}\small
\begin{aligned}
X_{t} = \begin{cases} X_{b} + Add(X_{hh}), \quad\text{if } N_{X_{hh}}<=N_{X_{b}}/2, \\
X_{b} + Add(X_{hh})[0:N_{X_{b}}/2],\quad\text{otherwise}\end{cases}
\end{aligned}
\end{equation}
where $X_{t}=36,000$ is the final number of sampled points, $X_{hh}$ are the semantic points corresponding to hands and face and $Add$ is defined as an $N_K$-steps recursion addition: $X_{hh}=X_{hh}+Add(X_{hh}+\mathcal{N}(0, \sigma_{HR}))$ with $\sigma_{HR}=0.07$ to sample closer to the surface where fine details lie while $\sigma_{LR}=5$.
As shown in~\cref{fig:marco_sampling}, the hands and face regions contain more points and fine details can be represented in those regions by the learned implicit function.
As sampling is done as a pre-processing step during training, this strategy is more tractable in terms of training time compared to training separate networks for each body region, whose merging is non-trivial.
For the depth-supervision strategy, we select $N_{pc}=15000$ sub-points of the input point clouds. See supplementary for additional details about the implementation.
\subsection{Ablation Studies}
\label{ssec:abl}
\begin{table}[t]
\vspace{-3mm}
\centering
\renewcommand\arraystretch{1.1}
\setlength{\tabcolsep}{7pt}
{\resizebox{\columnwidth}{!}{
\begin{tabular}{@{}l|ccc|ccc@{}}
\toprule
\multirow{2}{*}{\textbf{Design}}    & \multicolumn{3}{c|}{\textbf{RenderPeople}} & \multicolumn{3}{c}{\textbf{THuman2.0}} \\ \cline{2-7} 
           & CD           & Normal        & P2S         & CD          & Normal      & P2S        \\ \hline
2D LR only &   2.653      &        .5611       &  2.682    &    1.998    &     .3734       &    1.939   \\
2D HR only &   2.170      &         .5131      &    1.973     &  1.147     &     .3074        &   1.008 \\
3D only    &   3.021      &       .6302        &  2.715      &      1.961  &     .5179        &    1.631   \\
w. rand. feat. &  2.136      &        .5081       &   1.947    &     1.052   &      .2823      &  0.954  \\
ANIM (Ours)   &   \textbf{2.075}     &        \textbf{.4990}       &  \textbf{1.936}      &       \textbf{0.913}         &    \textbf{.2545}         &       \textbf{0.878}  \\
\bottomrule
\end{tabular}
}}
\vspace{-3.5mm}
\caption{Quantitative results obtained by modifying the architecture of the network.}
\vspace{-6mm}
\label{tab:des}
\end{table}

\noindent \textbf{Network architecture}. To prove the effectiveness of the proposed network architecture, we modify it as follows:
\\\textbullet\, \textbf{2D feature only}: Only RGB and normals are exploited by either the HR-FE  (2D HR only) or the LR-FE (2D LR only) networks. The VFE is not implemented.
\\\textbullet\, \textbf{3D feature only}: The final estimation is obtained by processing only the depth map. HR-FE and LR-FE are not implemented and RGB and surface normals are not used.
\\\textbullet\, \textbf{w. random feature}: Similar to previous work~\cite{peng2020neural}, we link random features to the voxel as input to the SparseConvNet instead of using the LR feature. This shows whether linking RGB features  retrieved from the input 2D image and surface normals to the voxels improves the performance.
\\Our proposed configuration obtains the lowest errors (see~\cref{tab:des}). The highest errors are obtained when either the 3D or the 2D encoders are not used, proving the effectiveness of using them together. Linking the LR feature to the voxel for the VFE improves \name performance.
\\\noindent \textbf{High quality details}. Our framework can reproduce significantly high-quality details on the final shape. We want to demonstrate the role of each component of the framework in learning details by setting up the following baselines:
\\\textbullet\, \textbf{w./o. normals}: The normals are not concatenated with the input RGB image and not considered in the reconstruction.
\\\textbullet\, \textbf{w./o. LR feature}: To show the importance of having a large receptive field, we test the approach without the LR-FE. The HR features are linked to the voxel as input to VFE.
\\\textbullet\, \textbf{w./o. HR feature}: The HR-FE is not implemented to demonstrate the effect of exploiting local HR features in addition to just normals and LR features. The output of LR-FE replaces the HR embedding of the original approach.
\\\textbullet\, \textbf{w./o. SA sampling}: To show how the semantic-aware sampling approach is essential to retrieve more details on the face and hands, we trained the approach without augmenting the sampling points on the face and hand regions.
\\\textbullet\, \textbf{w./o.} $\mathbf{L_{depth}}$: To demonstrate the effectiveness of the depth-supervision, we train \name with only the $L_{sdf}$ loss.
\\As shown in~\cref{tab:deta}, we conclude that each component in our proposed framework is fundamental to improving the quality of reconstruction results. We observed performance drops if any component is omitted, with the highest accuracy obtained when \name leverages all its components.  See supplementary for a qualitative evaluation.

\begin{table}[t]
\vspace{-3mm}
\centering
\renewcommand\arraystretch{1.1}
\setlength{\tabcolsep}{6pt}
{\resizebox{\columnwidth}{!}{
    \begin{tabular}{@{}l|ccc|ccc@{}}
    \toprule
    \multirow{2}{*}{\textbf{Design}} & \multicolumn{3}{c|}{\textbf{RenderPeople}} & \multicolumn{3}{c}{\textbf{THuman2.0}} \\ \cline{2-7} 
                    & CD           & Normal        & P2S         & CD          & Normal      & P2S        \\ \hline
    w/o normals     &    2.271     &    .5156           &    2.123    & 1.376      &   .3126          &  0.940    \\
    w/o LR feature &   2.453      &        .5611       &  2.282    &    1.653    &       .2954      &   2.554   \\
    w/o HR feature  &  2.605      &       .5320        & 3.176       &     2.649   &    .2599         &   3.323    \\
    w/o SA sampl.   &   2.636    &      .5328         &    2.238    &     0.993       &       .2710      &   0.908         \\
 w/o $L_{depth}$   &   \textbf{2.060}      &         .5647     &   1.956    &     0.947   &      .2689       &  0.915    \\
    ANIM (Ours)        &   2.075     &        \textbf{.4990}       &  \textbf{1.936}      &       \textbf{0.913}         &    \textbf{.2545}         &       \textbf{0.878}     \\
    \bottomrule
    \end{tabular}
}}
\vspace{-3.5mm}
\caption{Quantitative evaluation to demonstrate the influence of the adopted configuration to create fine details in the shapes.}
\label{tab:deta}
\vspace{-6mm}
\end{table}
\subsection{Comparisons to the State of the Art}
\label{ssec:comp}
\begin{figure}[t]
\vspace{-3mm}
\centering
\includegraphics[width=0.9\linewidth]{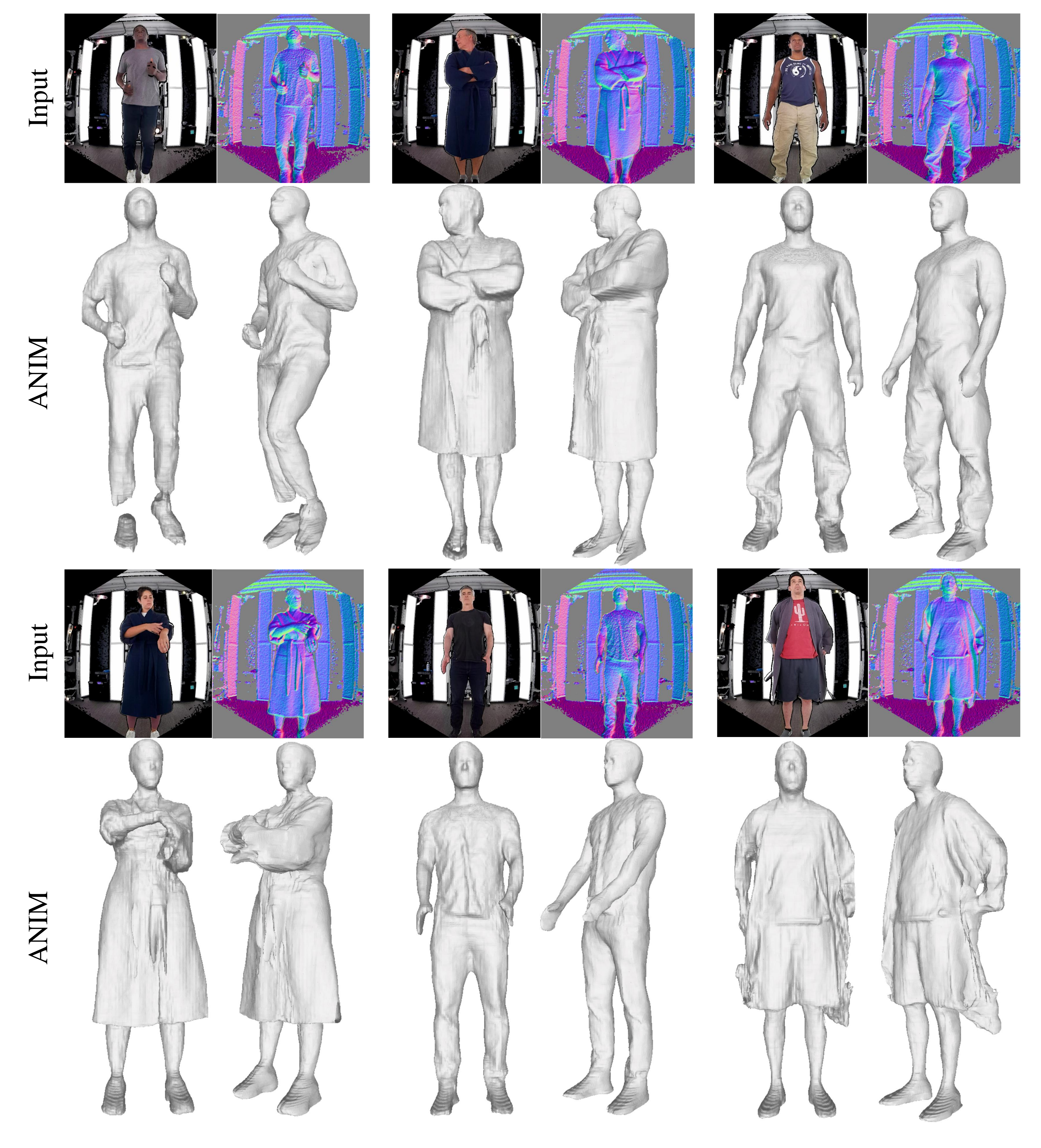}
\vspace{-4mm}
\caption{\name reconstructs fine-level cloth details such as wrinkles on the cloth and body with high accuracy even when the input is a consumer-grade RGB-D camera (Azure Kinect).}
\label{fig:kinect}
\vspace{-6.5mm}
\end{figure}

\begin{figure*}[ptb]
\centering
\includegraphics[width=0.95\linewidth]{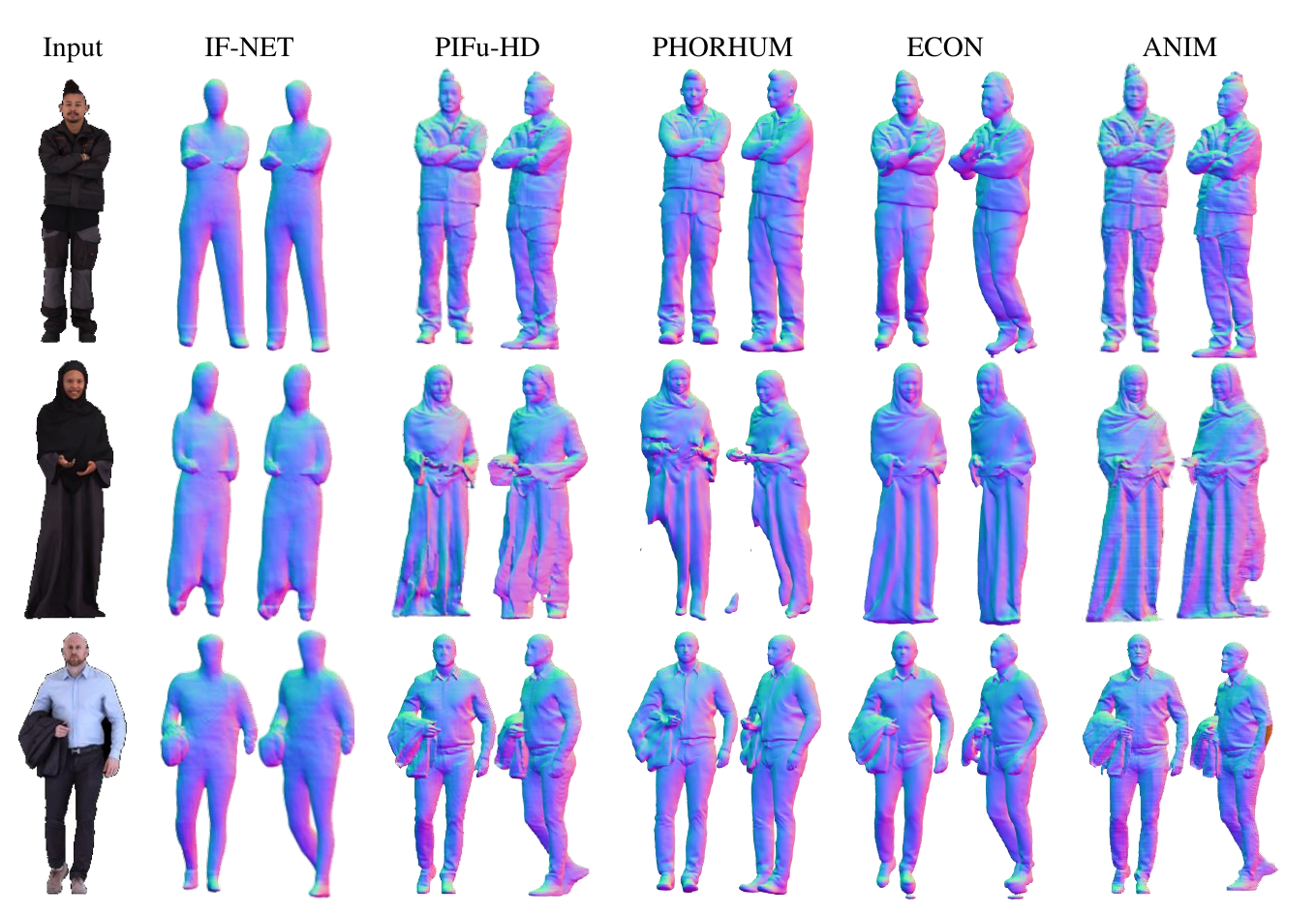} 
\vspace{-3mm}
\caption{Qualitative comparisons with state-of-the-art approaches on RenderPeople dataset given different kinds of input.}
\label{fig:qual}
\vspace{-2mm}
\end{figure*}

\begin{figure*}[ptb]
\centering
\includegraphics[width=0.95\linewidth]{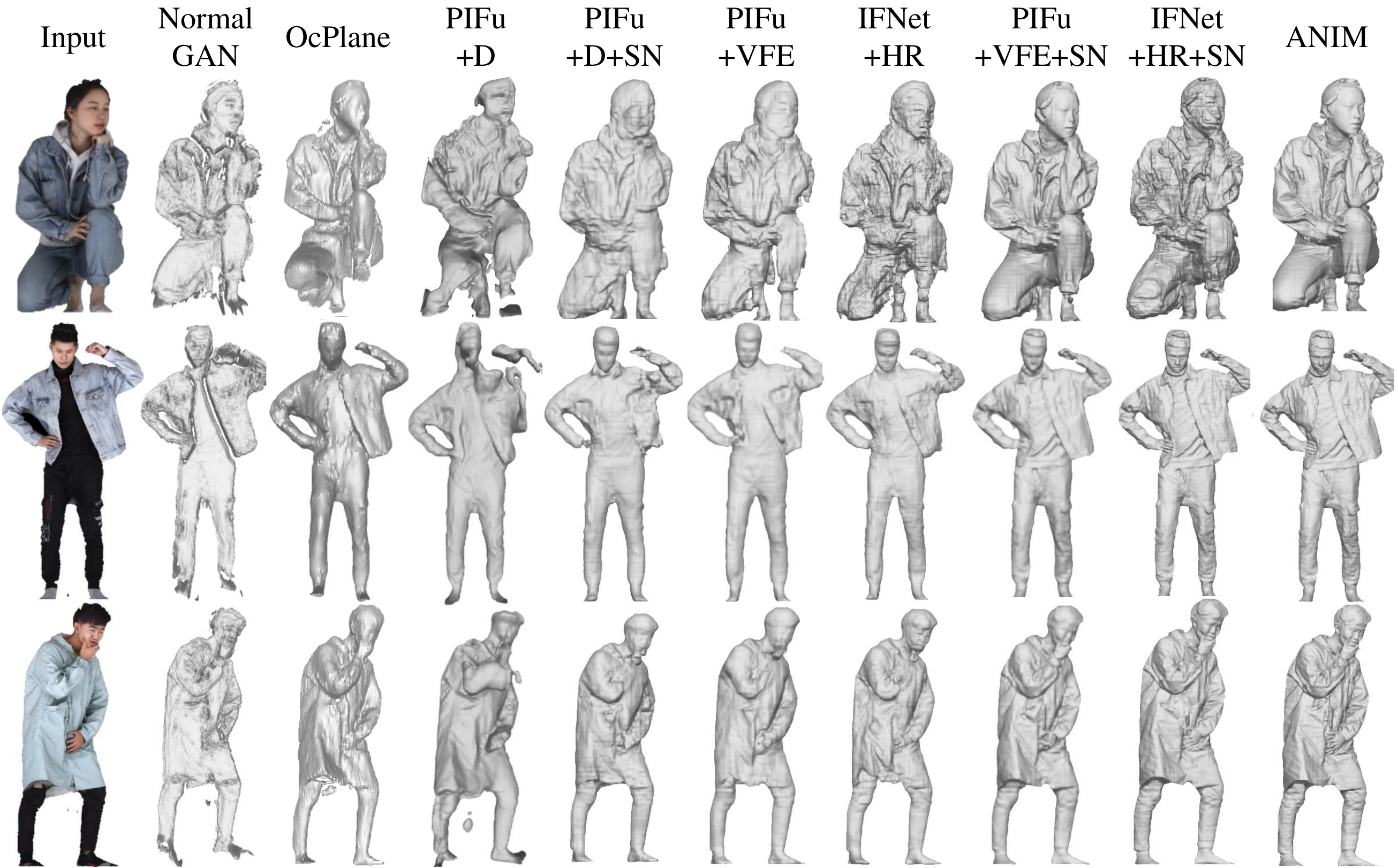} 
\vspace{-3mm}
\caption{Qualitative comparisons with state-of-the-art approaches on THuman2.0 dataset given an RGB-D image as input.}
\label{fig:qual_rgbd}
\vspace{-7mm}
\end{figure*}

Our goal is to demonstrate the advantages of using RGB-D data over other inputs by comparing \name with methods that rely on different single-input data, such as only RGB image (SuRS~\cite{SuRSECCV2022},PHORHUM~\cite{alldieck2022photorealistic}), surface normals and parametric models (PaMIR~\cite{Zerong2020PaMIR}, PIFuHD~\cite{saito2020pifuhd}, ICON~\cite{ICONCVPR2022}, ECON~\cite{xiu2023econ}) or point cloud (IF-Net~\cite{chibane20ifnet}). 
Additionally, we want to highlight the superiority of \name against related works that infer the 3D shape from a single RGB-D input (NormalGAN~\cite{wang2020normalgan}, OcPlanes~\cite{zhao2022occupancy}). 
We further adapt PIFu~\cite{PIFuICCV19} to RGBD-based reconstruction (PIFu+$D$) to establish fair comparisons, by concatenating depth with the RGB inputs. 
To demonstrate the effectiveness of the multi-resolution image extractor and of the VFE, we modify the architecture of PIFu and IF-Net by adding the VFE to PIFu and the HR-FE to IF-Net. 
PIFu+VFE processes the depth map with the VFE to extract geometric features, which are concatenated with the features of PIFu 2D encoder. 
IF-Net+HR processes RGB images with HR-FE and concatenates the feature with the output of the IF-Net 3D encoder. 
We repeat all aforementioned modifications by adding surface normals as input (indicated by + $S_N$). 
\\\noindent\textbf{Quantitative comparisons.} In ~\cref{tab:comp} we report a plethora of quantitative comparisons of \name against other works on the RenderPeople~\cite{renderpeople} and THuman2.0~\cite{tao2021function4d} datasets and showcase that our approach outperforms top performing competing methods by a large margin in both fidelity and accuracy. 
Extracting geometric information from the input depth map achieves better results compared to methods that estimate surface normals and parametric models. 
Moreover, the complete information extracted from the combination of RGB, normals and depth allows \name to outperform all the methods that rely on a single input. The novel architecture of \name guarantees the highest performance compared to methods that reconstruct 3D shapes from a single RGB-D. 
\begin{table}[t!]
\vspace{-3mm}
\centering
\renewcommand\arraystretch{1.1}
\setlength{\tabcolsep}{4pt}
{
    \resizebox{\columnwidth}{!}{
    \begin{tabular}{@{}l|l|ccc|ccc@{}}
    \toprule
    \multirow{2}{*}{ } & \multirow{2}{*}{\textbf{Methods}} & \multicolumn{3}{c|}{\textbf{RenderPeople}} & \multicolumn{3}{c}{\textbf{THuman2.0}} \\ 
                          &          & CD           & Normal        & P2S         & CD        & Normal        & P2S        \\ \hline

    \multicolumn{1}{c|}{\multirow{4}{*}[-4.5ex]{\rotatebox[origin=c]{90}{\textbf{\textit{Other input}}}}} &
    IF-Net~\cite{chibane20ifnet}                           &    4.546          &   .7732            &   4.375          &   1.924        &       .4181       &    1.847        \\
   \multicolumn{1}{l|}{}                  & PaMIR~\cite{Zerong2020PaMIR}                           &      3.944        &         .6562      &   3.261         &   2.602        &       .3721        &    2.727        \\
  \multicolumn{1}{l|}{}                  &  PIFuHD~\cite{saito2020pifuhd}                         &     2.415         &       .5495        &   2.381         &     3.625      &        .2730       &    3.462        \\
   \multicolumn{1}{l|}{}                  & ICON~\cite{ICONCVPR2022}                            &      2.330       &      .5886         &         2.301   &    2.093       &      .2791        &     1.112        \\
    
   \multicolumn{1}{l|}{}                  & PHORHUM~\cite{alldieck2022photorealistic}                            &      2.390       &      .5341         &         2.349   &    3.199       &      .2634        &     2.988        \\
    
   \multicolumn{1}{l|}{}                  & ECON~\cite{xiu2023econ}                            &      2.261       &      .5536         &         2.269   &    1.339       &      .2736        &     1.412        \\
    
   \multicolumn{1}{l|}{}                  & SuRS~\cite{SuRSECCV2022}                            &      2.810       &      .5909         &         2.884   &    1.290       &      .2945        &     1.695        \\\hline
    \multirow{12}{*}[4.5ex]{\rotatebox[origin=b]{90}{\textbf{\textit{Single RGB-D input}}}} &
    PIFu+$D$                        &      4.650      &       .7567        &         4.314    &     4.441      &       .3704        &     3.835          \\
  \multicolumn{1}{l|}{}                  &  PIFu+$D$+$S_N$                  &     3.544       &          .6754     &      3.653       &      4.379     &         .4302      &   3.983         \\
  \multicolumn{1}{l|}{}                  &  PIFu+VFE                        &   2.718           &      .5747         &  2.089           &    4.444      &       .3240        &     3.834       \\
 \multicolumn{1}{l|}{}                  &   PIFu+VFE+$S_N$                  &     2.242          &      .5218        &          2.076    &    0.924       &        .2548       &     0.880       \\
 \multicolumn{1}{l|}{}                  &   IF-Net+HR                    &        2.352      &       .5304        &     1.962       &   1.403        &       .3754        &      1.322     \\
  \multicolumn{1}{l|}{}                  &  IF-Net+HR+$S_N$\!            &  2.164          &    .4995           &      1.953       &  1.079        &        .2875       &     0.993       \\
  \multicolumn{1}{l|}{}                  &  NormalGAN~\cite{wang2020normalgan}                    &        3.924      &       .7912        &     3.224       &   2.830        &       .5914        &      2.658      \\
 \multicolumn{1}{l|}{}                  &   OcPlane~\cite{zhao2022occupancy}            &  5.619           &    .5324           &      4.188       &  3.734         &        .3303       &     3.728       \\
    
  \multicolumn{1}{l|}{}                  &  ANIM (Ours)             &   \textbf{2.075}     &        \textbf{.4990}       &  \textbf{1.936}      &       \textbf{0.913}         &    \textbf{.2545}         &       \textbf{0.878}        \\
    \bottomrule
    \end{tabular}
}
}
\vspace{-3.5mm}
\caption{Quantitative comparisons with state-of-the-art approaches in 3D human reconstruction from a single input.}
\vspace{-7mm}
\label{tab:comp}
\end{table}
\\\noindent\textbf{Qualitative comparisons} with works that reconstruct the 3D shape from input different than RGB-D for RenderPeople~\cite{renderpeople} are shown in~\cref{fig:qual}, whilst~\cref{fig:qual_rgbd} shows reconstruction from RGB-D for THuman2.0~\cite{tao2021function4d}. 
The quality of the 3D shapes reconstructed by \name is significantly higher than all the other methods. 
\name outperforms methods that do not process RGB-D data because the integration of depth information, alongside RGB, is essential for achieving high-quality and accurate estimations. 
By learning spatial relationships from the geometric information extracted from the depth maps, \name effectively avoids depth ambiguity, resulting in more accurate reconstructions compared to methods that rely on estimating parametric models. 
The multi-resolution feature extractors employed by \name ensure the reproduction of finer details in contrast to other approaches. 
RGB-D methods are outperformed thanks to the depth supervision strategy and the combination of 2D multi-resolution and 3D geometric features, leveraging pixel-voxel-aligned properties inherent in the implicit representation. 
Implementing the introduced modules within benchmark works (PIFu, IF-Net) significantly improves their quantitative and qualitative performance, but the highest accuracy is still obtained by \name.
\\\noindent\textbf{Real-world capture.}
We provide qualitative results of 3D human shape reconstructed by \name after being fine-tuned with the \datasetname dataset (\cref{fig:kinect}).
The RGB-D input acquired with the Azure Kinect exhibits significant noise in the depth maps and surface normals. 
However, thanks to the fine-tuning, \name faithfully reconstructs the geometry of the clothed human, capturing fine-level details such as wrinkles and specific body features like the face and hands. See supplementary materials for additional results.

\vspace{-1mm}
\section{Conclusion}
\label{sec:conclusion}
\vspace{-1mm}

We introduce \name, a novel neural implicit model that reconstructs accurate and high-fidelity 3D humans from single RGB-D images, outperforming existing methods over other kinds of input and proving the benefit of leveraging RGB-D data.  
We demonstrate the effectiveness of combining both multi-resolution pixel-voxel-aligned features and a novel depth-supervision strategy to address depth ambiguity and reconstruct high-quality 3D human shapes.
We also present the multi-modal dataset \datasetname consisting of high-quality 3D scans and real-world captures, obtained with a high-resolution camera system paired with a consumer-grade RGB-D camera. \datasetname significantly leverages \name for human reconstruction, and can be valuable to the community for neural implicit 3D human reconstruction.
Future work includes exploring temporal information for fusion of body pose and appearance across time.
\\
{\footnotesize \textbf{Acknowledgement}: This work was supported by Meta, UKRI EPSRC and BBC Prosperity Partnership AI4ME: Future Personalised Object-Based Media Experiences Delivered at Scale Anywhere EP/V038087.}
{
\small
\bibliographystyle{ieeenat_fullname}
\bibliography{main}
}
\clearpage
\setcounter{page}{1}
\maketitlesupplementary
\section{Overview}
In this supplementary material we  provide:
 \begin{enumerate}
     \item Additional details about the implementation of \name
     \item Details about the architectural design of VFE (Voxel Feature Extractor)
     \item Further details on \datasetname 
     \item Limitations of the proposed approach
     \item Additional results obtained by applying \name on real noisy data captured with Azure Kinect
     \item Qualitative results for the ablations studies presented in the main paper
     \item Additional qualitative results, to further demonstrate that \name and the new technical contributions we propose clearly outperform prior works on reconstruction quality.
 \end{enumerate}
\section{Implementation details}
In our proposed network architecture, the normals and the RGB images are concatenated and processed by the two hourglass architectures with four stacks each: the HR-FE outputs an embedding of resolution $256\times256\times256$ while the resolution of the features obtained from LR-FE is $256\times128\times128$. 
The former are bi-linearly interpolated with the ground-truth points projected on the input image to align the point and the feature in the 2D space. 
The latter are given as input to the VFE along with a voxel created from the input depth map. 3D points from the depth map are obtained by transforming 2D image coordinates to 3D world coordinates using the camera parameters, prior to normalization. The voxel is created from these 3D points. The LR features are aligned with the voxel, which is created with as many voxels as the number of channels of the LR feature ($256$). The VFE is a novel SparseConvNet U-Net style architecture, 
based on SparseConvNet~\cite{SparseConvNet18} that has shown to be efficient for the task of 3D object detection when the input is sparse. 
Following \cite{peng2020neural}, 
for any point in 3D space, we tri-linearly interpolate the latent codes from multi-scale code volumes with the ground truth point. 
The VFE and the HR-FE features are concatenated and finally classified by the MLP with a number of neurons equal to (369, 512, 256, 128, 1). 
The same features extracted from the VFE and the HR-FE are then interpolated with the point cloud for the depth-supervision.
We implement our proposed framework using PyTorch and run training and testing with NVIDIA Tesla V100 GPUs. We train the neural networks with Adam optimizer and a learning rate $lr=1e-4$ and $\delta=1.25$.
Inference time for one image, without code optimization, is in the order of the second.
\\For the comparisons in~\cref{ssec:comp} IF-Net, PaMIR, ICON, SuRS, OcPlans, and (6) PIFu and IF-Net 
variants are retrained with the same dataset and configuration as \name. PIFuHD, ECON, PHORHUM and NormalGAN are not retrained due to unavailability of training code. We used their checkpoints for evaluation. All methods are tested on the same datasets (RenderPeople~\cite{renderpeople}, THuman2.0~\cite{tao2021function4d}).
\\\noindent \textbf{Ethical concerns}. \name was trained on public datasets that do not reveal the identity of subjects. \name aims at faithfully capturing full-body humans without alteration and body distortion, avoiding potential misuse or misrepresentation.
\section{VFE Architecture}
\begin{table*}[t!]
\renewcommand{\tabcolsep}{8pt}
    \centering
    \begin{tabular}{@{\hspace{0.7mm}}c@{\hspace{0.7mm}}c|@{\hspace{0.7mm}}c|@{\hspace{0.7mm}}c@{\hspace{0.7mm}}c}
    \hline
    & Layer& Layer Description & Output Dimension \\ \hline
    & & Input volume & \(D\times H\times W\times256\) \\ \hline
    & 1-3 & (3$\times$3$\times$3 conv, 16 features, stride 1) $\times$2
    &D$\times$H$\times$W$\times$16\\
    &4 &(3$\times$3$\times$3 conv, 32 features, stride 2)
    &1/2D$\times$1/2H$\times$1/2W$\times$32\\
    &5-6  & (3$\times$3$\times$3 conv, 32 features, stride 1) $\times$2
   &1/2D$\times$1/2H$\times$1/2W$\times$32\\
    &7 & (3$\times$3$\times$3 conv, 64 features, stride 2)
    &1/4D$\times$1/4H$\times$1/4W$\times$64\\
    &8-10&(3$\times$3$\times$3 conv, 64 features, stride 1) $\times$ 3
    &1/4D$\times$1/4H$\times$1/4W$\times$64\\
    &11&(3$\times$3$\times$3 conv, 128 features, stride 2)
    &1/8D$\times$1/8H$\times$1/8W$\times$128\\
    &12-15&(3$\times$3$\times$3 conv, 128 features, stride 1)$\times$4
    &1/8D$\times$1/8H$\times$1/8W$\times$128\\
    &16 &3$\times$3$\times$3 invConv, 64 features, stride 1
    
    &1/4D$\times$1/4H$\times$1/4W$\times$64\\
    &-& concat output 16/10&1/4D$\times$1/4H$\times$1/4W$\times$128\\
    &17 &3$\times$3$\times$3 conv, 32 features, stride 1
    &1/4D$\times$1/4H$\times$1/4W$\times$64\\
&18-20&(3$\times$3$\times$3 conv, 32 features, stride 1) $\times$ 3
    &1/4D$\times$1/4H$\times$1/4W$\times$64\\
    &21 &3$\times$3$\times$3 invConv, 32 features, stride 1
    
    &1/2D$\times$1/2H$\times$1/2W$\times$32\\
    &-& concat output 21/6& 1/2D$\times$1/2H$\times$1/2W$\times$64\\
    &22&3$\times$3$\times$3 conv, 32 features, stride 1
    &1/2D$\times$1/2H$\times$1/2W$\times$32\\
&23-24&(3$\times$3$\times$3 conv, 32 features, stride 1) $\times$ 2
    &1/2D$\times$1/2H$\times$1/2W$\times$32\\
    &25&3$\times$3$\times$3 invConv, 16 features, stride 1
    &D$\times$H$\times$W$\times$16\\
    &-& concat output 25/3&D$\times$H$\times$W$\times$32\\
    &26 &3$\times$3$\times$3 conv, 16 features, stride 1
    &D$\times$H$\times$W$\times$16\\
&27-28&(3$\times$3$\times$3 conv, 16 features, stride 1) $\times$ 2
    &D$\times$H$\times$W$\times$16\\
    \hline
    \end{tabular}
    \caption{VFE SparseConvNet U-net Architecture.}
    \label{tab:vfe}
\end{table*}
We report in~\cref{tab:vfe} the detailed architecture of VFE, which consists of a SparseConvNet U-net that we designed for ANIM. The SparseConvNet implements spatially sparse convolutional networks~\cite{SparseConvNet18}. The VFE architecture is implemented using sub-manifold sparse convolution operations. The table gives the sizes of the different layers and of the receptive fields. We experimented with various variants and report the ones that returned the best results in our experiments.

\section{\datasetname dataset details}
As explained in~\cref{sec:dataset} of the main paper, the performance of neural implicit models significantly deteriorates when tested with raw data from consumer-grade sensors due to the severe input noise. To address this problem, we curated a new dataset (\datasetname) consisting of RGB-D noisy data captured with Azure Kinect and high-quality 3D ground-truth meshes reconstructed using a high-resolution camera system that employs active stereo and multi-view cameras~\cite{3dmd}. We fine-tune \name with this dataset to reconstruct accurate and high-quality 3D human shapes from real-world data, mitigating the impact of the sensor noise. This section provides further details on the system used for data capture and presents examples of data of \datasetname.
\\The capture system comprises two subsystems, with 1 Azure Kinect camera and 32 multi-view stereo cameras from~\cite{3dmd}.
To acquire the data, we calibrate the two systems in order to align the 3D ground-truth meshes with the RGB-D data. The collected dataset consists of 
31 subjects, with 16 women and 15 men captured, each subject performing a set of scripted animations (\eg, standing, walking, turning, jogging, stretching, putting on/taking off clothes). Some examples of data are shown in Fig.~\ref{fig:supp_results_azure}.

Datasets that integrate high-resolution 3D ground-truth shapes with raw RGB-D data are currently unavailable. The introduction of \datasetname is a valuable contribution to the research community in the context of neural implicit 3D human reconstruction. This dataset helps to mitigate domain gaps, providing researchers with a resource that facilitates the development of effective techniques in this domain.

\section{Limitations}
\begin{figure}[t]
\centering
\includegraphics[width=1.0\linewidth]{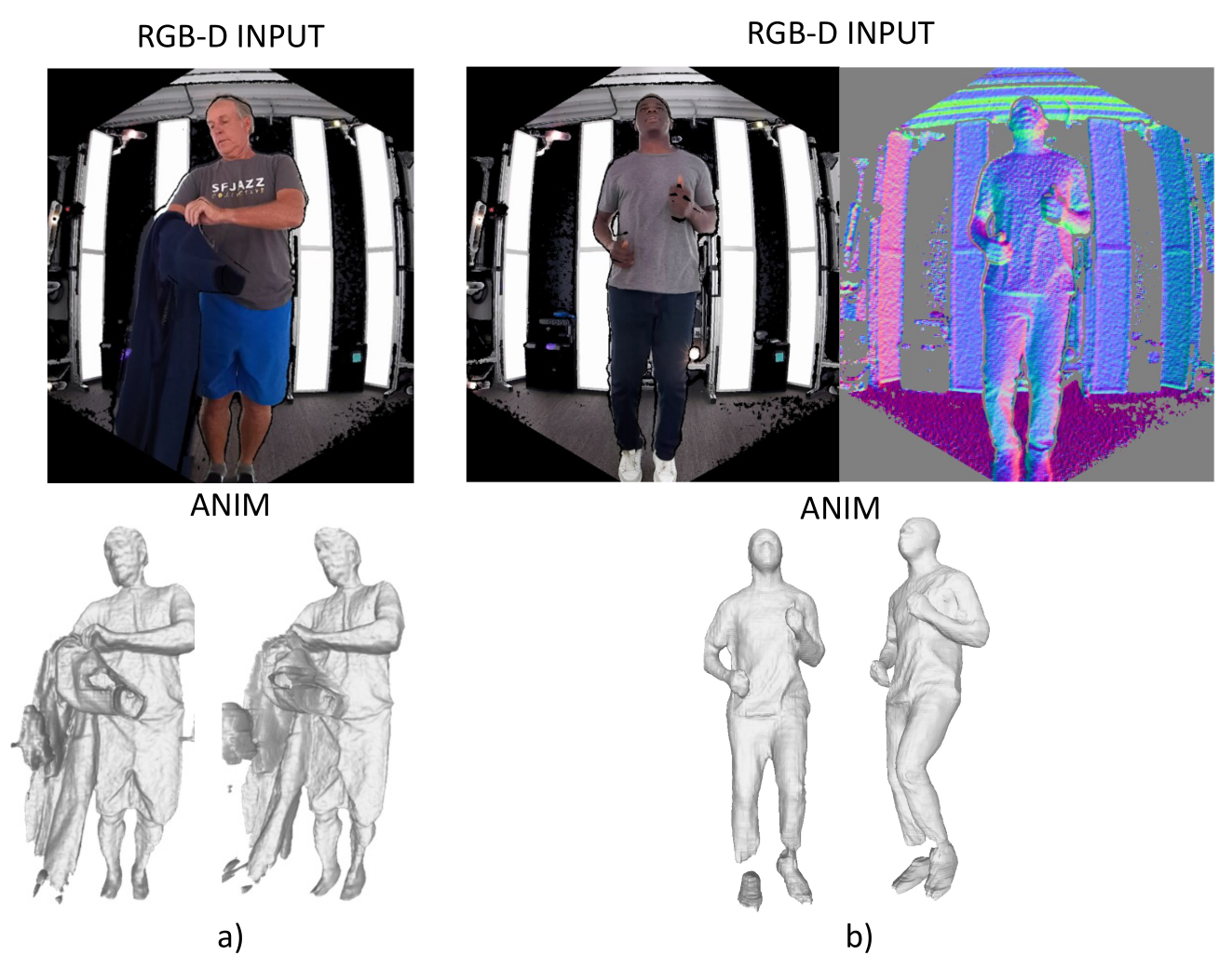}
\caption{Limitations of \name. Accuracy is reduced in challenging scenes (a). Noise still affects the final reconstruction is some body parts of the shape (b).}
\label{fig:limitation}
\end{figure}

Failure cases can arise from challenging scenes that include arbitrary objects or complex motions (\eg taking of clothes) as shown in~\cref{fig:limitation}a. 
\\The accuracy of \name applied to real-world data is slightly lower than the one achieved with synthetic data since \name is still influenced by the noise of the input raw data, which can affect the reconstruction as shown in~\cref{fig:limitation}b where the ankle of the model is not reconstructed.
\\The model could be further fine-tuned to learn specific sensor noise and mitigate domain gaps.
\section{Additional results on real-world data}
\begin{figure*}[t]
\centering
\includegraphics[width=\textwidth]{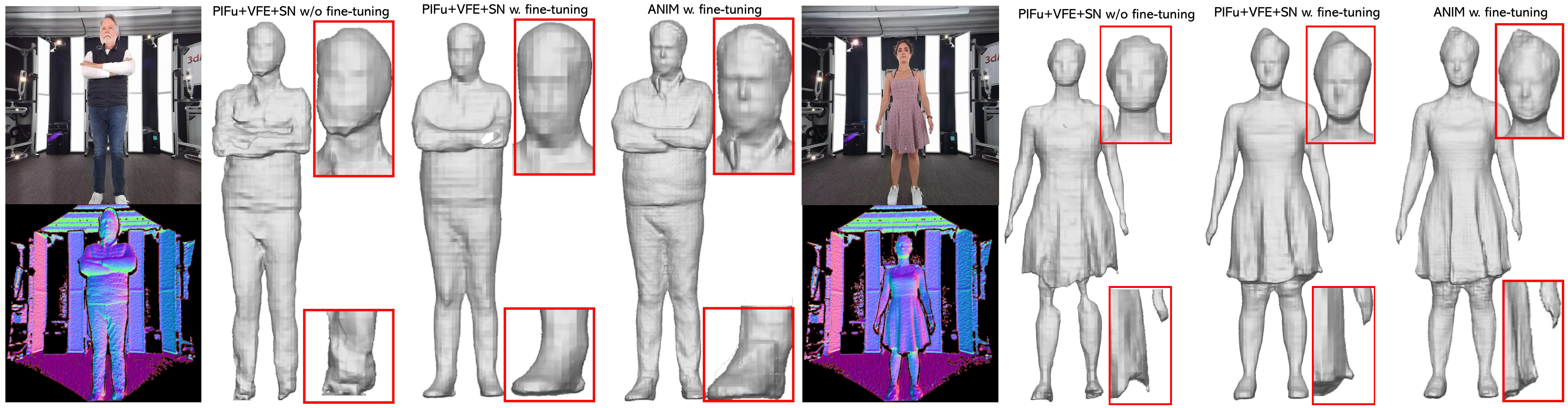}
\caption{\label{fig:supp_qual_comp_pifu} Qualitative comparison with PIFu+VFE+SN using real-world data.}
\end{figure*}
\begin{figure*}
\centering
\includegraphics[width=\textwidth]{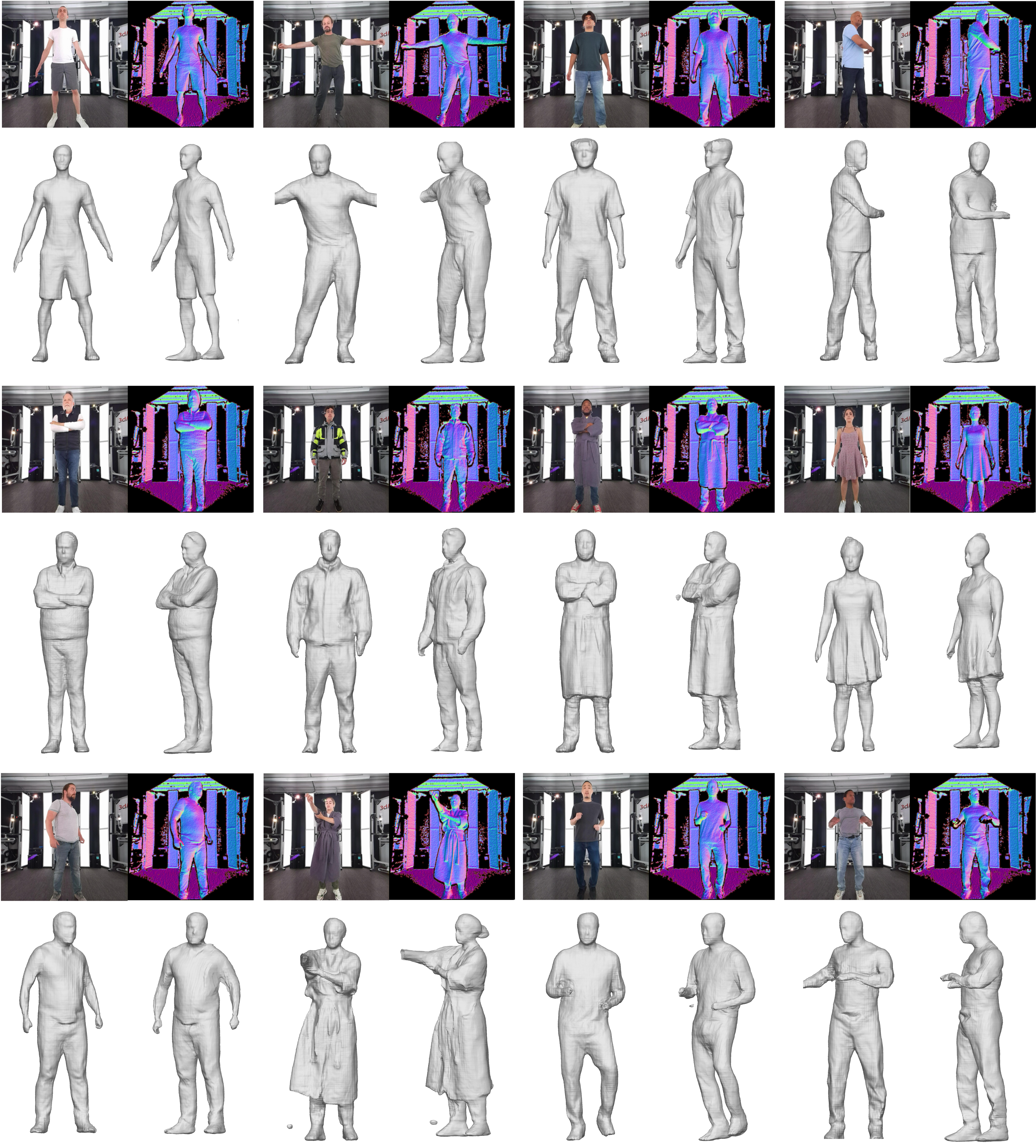}
\caption{\label{fig:supp_results_azure} More reconstruction results by ANIM using a consumer-grade RGB-D camera (Azure Kinect) as an input. ANIM is capable of handling various human body and cloth typologies ranging from a skirt to a bath robe and is agnostic to diverse human poses.}
\end{figure*}

We test \name on real-world data obtained with an Azure Kinect after fine-tuning with additional 16k frames, consisting of around 800 frames in average from a single view of 21 subjects. ~\cref{fig:supp_results_azure} shows examples of \name reconstruction from real single RGB-D images captured with a Kinect Azure.
Our approach can retrieve high-quality details on the final mesh even if the input normals and depth are noisy. \name can eliminate the noise of the consumer-grade sensor, significantly improving the reconstruction with accurate and high-quality 3D human shapes.
We present further qualitative comparisons among different methods using real-world data.
Given the inherent challenges associated with the real-world dataset, we show results from one of the most competitive methods, PIFu+VFE+SN, both before and after finetuning.
As illustrated in Figure~\ref{fig:supp_qual_comp_pifu}, finetuning PIFu+VFE+SN on ANIM-Real yeilds qualitative improvements, yet not on par with ANIM.
\section{Ablation Studies}
\begin{figure*}[t]
\centering
\includegraphics[width=0.85\linewidth]{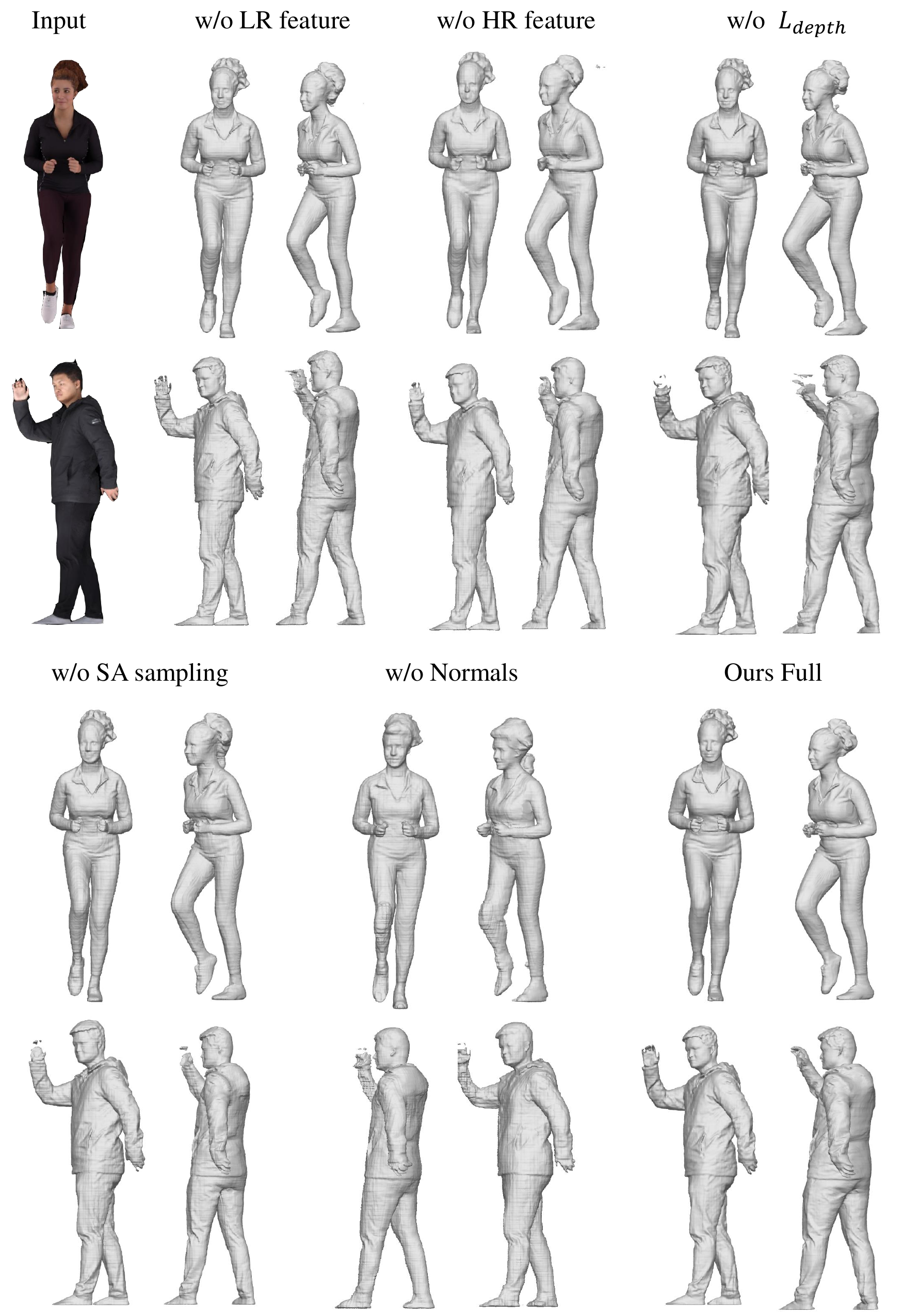}
\vspace{-3mm}
\caption{We conducted an ablation study on the components of \name that influence the reconstruction quality. We show reconstructions of $2$ subjects (one from THuman2.0~\cite{tao2021function4d} and the other from RenderPeople~\cite{renderpeople}) captured by a single-view RGBD image (\ie partial view), from frontal and 45-deg side views. Our full \name model provides high-quality reconstructions with facial expressions, hands, and cloth wrinkles with fine-level details, without shape distortion along the camera view. Please zoom in the figure to better see details.}
\label{fig:ablation1}
\end{figure*}

\begin{figure*}[t]
\centering
\includegraphics[width=0.85\linewidth]{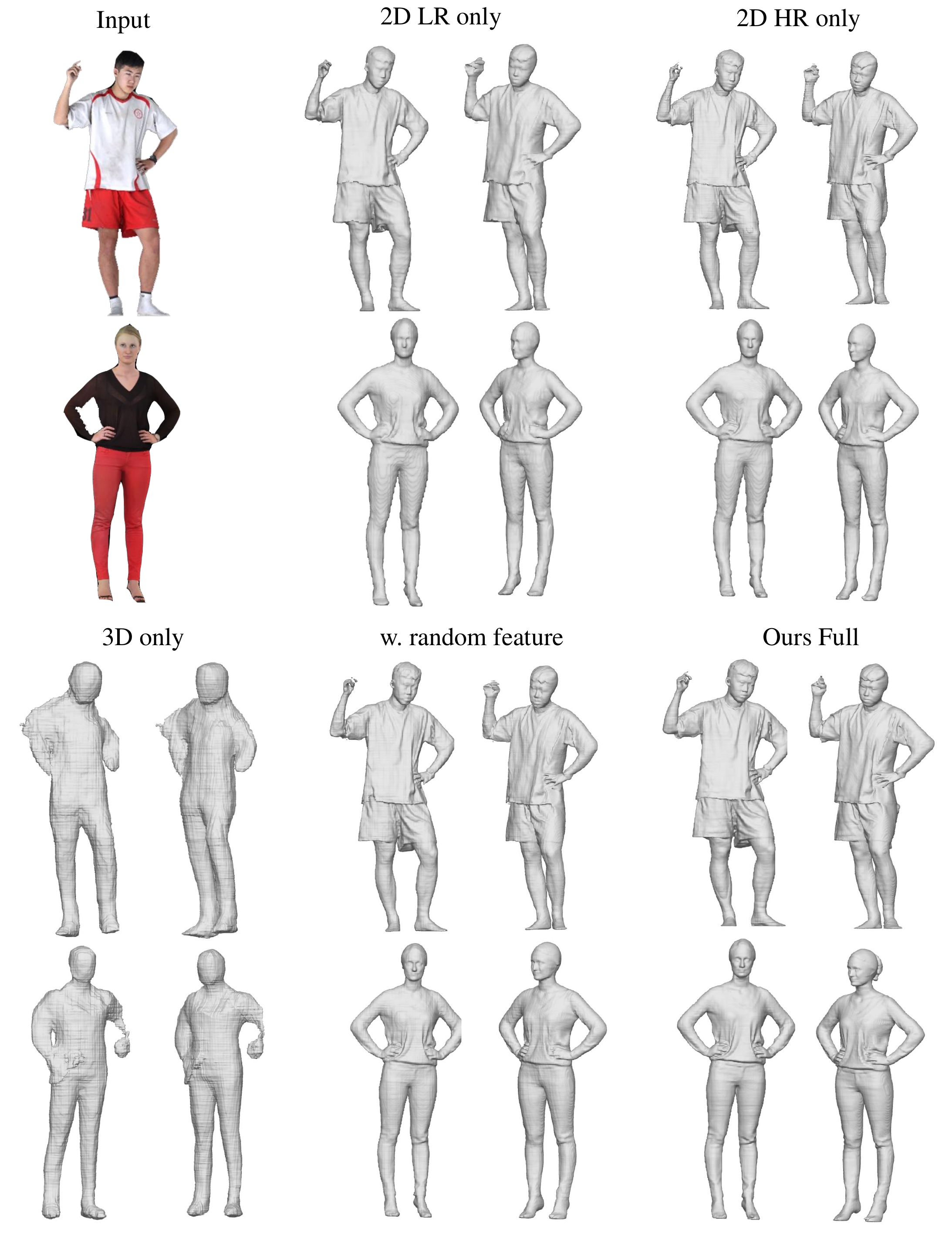}
\caption{We conducted an ablation study where components of \name were removed one-by-one to prove the superiority of the proposed architecture. We show reconstructions of $2$ subjects (one from THuman2.0~\cite{tao2021function4d} and the other from RenderPeople~\cite{renderpeople}) captured by a single-view RGBD image (\ie partial view), from frontal and 45-deg side views, with colored normals. Our full \name model provides more accurate results. Please zoom in the figure to better see details.}
\label{fig:ablation2}
\end{figure*}

We illustrate qualitative comparisons for the ablation studies presented in~\cref{ssec:abl} of the main paper.
The labels used in the figures are consistent with the ablation study conducted in~\cref{tab:des} and~\cref{tab:deta} in the main paper.~\cref{fig:ablation1} illustrates the role that each module of \name plays in representing high-quality details in the final reconstruction, with the highest-quality shapes obtained when all the modules are exploited. More specifically, fewer details are represented in the face and hands of the model when spatial-aware sampling is not applied. The importance of normals and HR feature can also be noticed by the reduced amount of details in the final reconstruction. Less accurate shapes are then obtained if LR feature is  not used. The introduction of depth supervision further increases the accuracy and the details in the reconstructed shapes.
Fig~\ref{fig:ablation2} demonstrates the effectiveness of the architecture of \name. Each key component was tested one-by-one and it is proved that the complete model outperforms the others with more accurate and highly-detailed 3D shapes.
\section{Additional qualitative Results}
\begin{figure*}[t]
\centering
\includegraphics[width=0.81\linewidth]{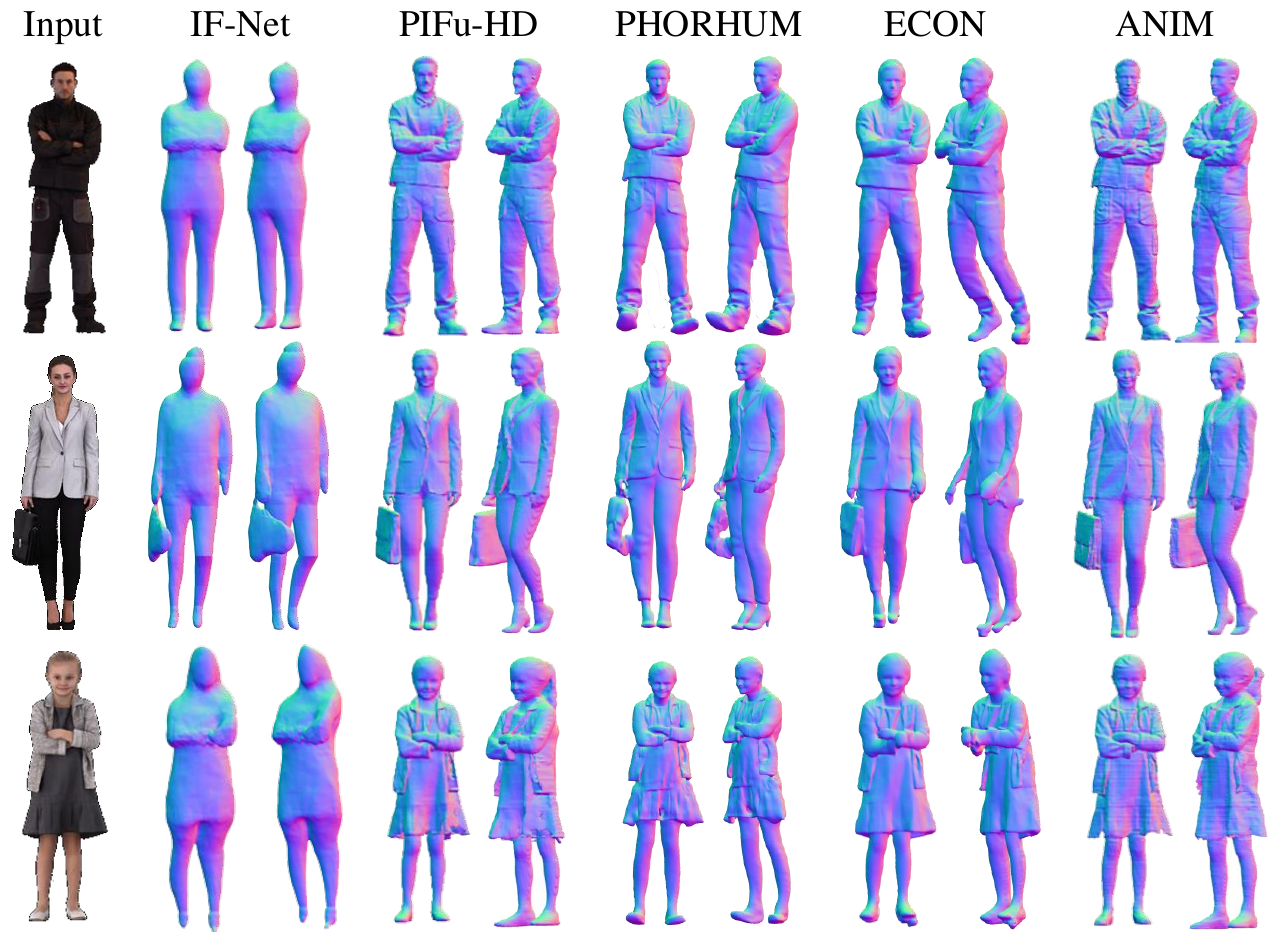}
\vspace{-3mm}
\caption{Additional comparisons with approaches that use single RGB image or partial point clouds as input. Data from RenderPeople \cite{renderpeople}. \name reconstructs full-body models with high accuracy, with cloth wrinkles, face and hand details, and without depth ambiguity (\ie distortion along camera view).}
\label{fig:main2}
\vspace{-3mm}
\end{figure*}

\begin{figure*}[t]
\centering
\includegraphics[width=0.85\linewidth]{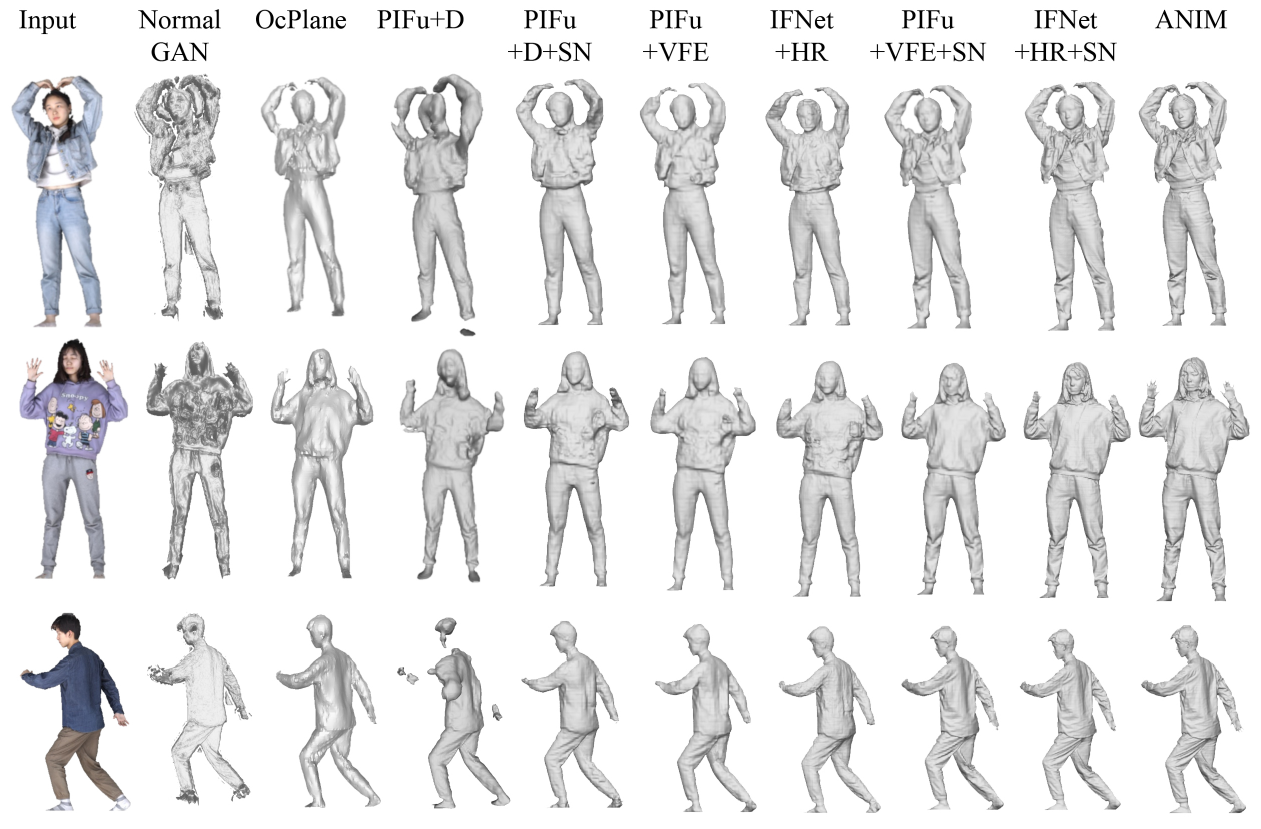}
\vspace{-3mm}
\caption{Additional comparisons with methods that use a single RGB-D image as input. Our core contributions can leverage state-of-the-art models, but only our complete \name model design returns the best results. We show reconstruction from the front view. Data from THuman2.0~\cite{tao2021function4d}. 
}
\label{fig:thuman2}
\end{figure*}

\begin{figure*}[ptb]
\centering
\includegraphics[width=0.8\textwidth]{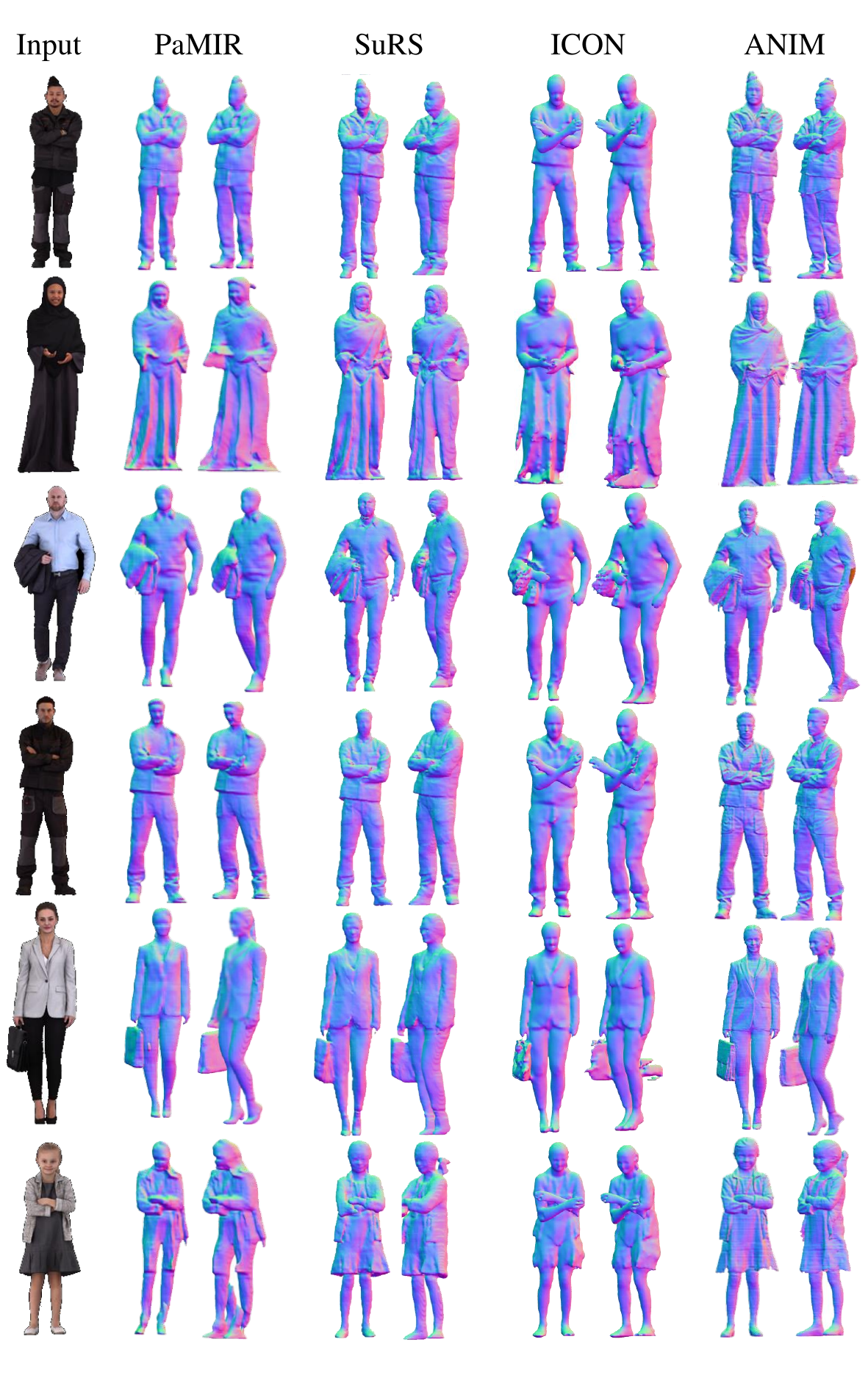}
\caption{Qualitative comparisons with approaches not illustrated in the main paper that use a single RGB image or partial point clouds as input. Data from RenderPeople \cite{renderpeople}.}
\label{fig:main_add}
\end{figure*}

\begin{figure*}[t]
\centering
\includegraphics[width=\linewidth]{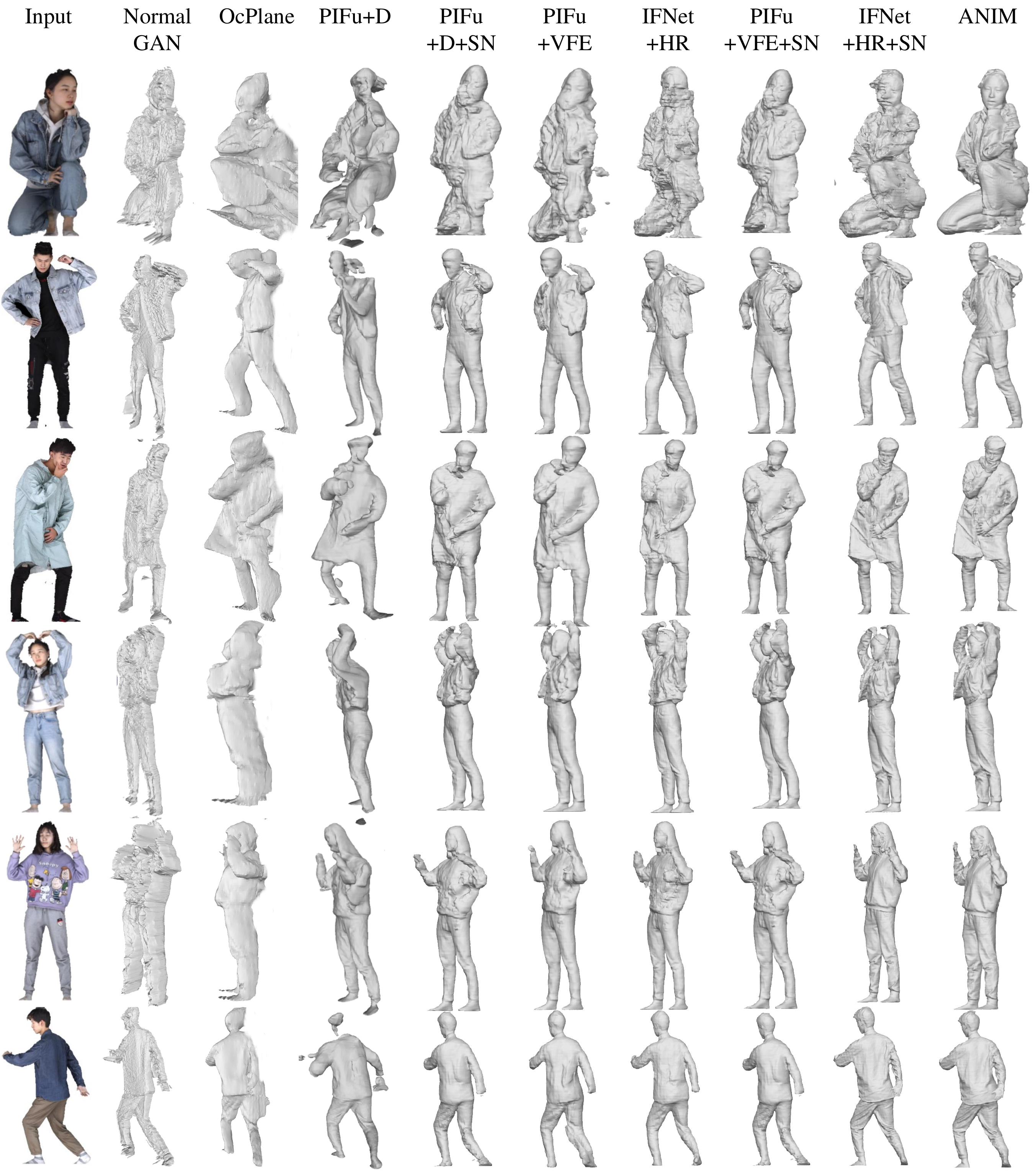}
\caption{Side-views of the 3D shapes reconstructed from an input RGB-D data showed in~\cref{fig:qual_rgbd} and~\cref{fig:thuman2}.
}
\label{fig:thuman2_side}
\end{figure*}

Additional qualitative comparisons for approaches that reconstruct the 3D shape from an input different than RGB-D are presented in \cref{fig:main2} while~\cref{fig:thuman2} shows additional results obtained by reconstructing 3D shapes from RGB-D data.
\name consistently generates high-fidelity reconstructions, with cloth wrinkles and high-quality faces and hands in accordance with the input RGB images thanks to our depth-supervision strategy. Depth ambiguity issues are also solved by leveraging the depth channel of the input data. Moreover, it is shown how the contributions we propose, such as using the VFE and the multi-resolution features of HR-FE and LR-FE, can be used to improve other approaches, but only our complete \name model design returns the best results.
\\\cref{fig:main_add} shows results of reconstructing 3D shapes from input different than RGB-D for other related methods that are not shown in the paper.
\\\cref{fig:thuman2_side} show the the side-view reconstruction of the results showed in~\cref{fig:qual_rgbd} and~\cref{fig:thuman2}. 


\end{document}